\documentstyle[jair,twoside,11pt,theapa,url]{article}
\input epsf 

\jairheading{29}{2007}{353--389}{9/06}{8/07}

\ShortHeadings{Logic Programs with Arbitrary Abstract Constraint Atoms}
{Son, Pontelli, \& Tu}
\firstpageno{353}

\long\def\COMMENT#1\ENDCOMMENT{\message{(Commented text...)}\par}

\date{}

\newtheorem{example}{Example}

\newtheorem{proposition}{Proposition}
\newtheorem{lemma}{Lemma}

\newtheorem{definition}{Definition}

\newtheorem{corollary}{Corollary}
\newtheorem{observation}{Observation}

\def\la{\leftarrow}
\def\naf{{\; not \;}}
\def\catom{{\textnormal{\em c-atom}}}
\def\qed{\hfill $\Box$}

\begin{document}

\title{
Answer Sets for Logic Programs with Arbitrary Abstract Constraint Atoms
}

\author{\name Tran Cao Son \email tson@cs.nmsu.edu\\
\name Enrico Pontelli \email epontell@cs.nmsu.edu\\
\name Phan Huy Tu  \email tphan@cs.nmsu.edu\\
\addr Computer Science Department \\
New Mexico State University  \\
Las Cruces, NM 88003, USA 
}

\maketitle
\begin{abstract}
In this paper, we present two alternative approaches to defining 
answer sets for logic programs with {\em arbitrary} types of 
abstract constraint atoms (c-atoms). 
These approaches generalize the fixpoint-based and the level mapping based 
answer set semantics of normal logic programs to the case of
logic programs with arbitrary types of c-atoms. The results are 
four different answer set definitions which are equivalent when 
applied to normal logic programs.  

The standard fixpoint-based semantics of logic programs
is generalized in two directions, called
answer set by reduct and answer set by 
complement. These definitions, which 
differ from each other in the treatment of negation-as-failure (\emph{naf})
atoms, make use of an immediate consequence operator to perform answer set 
checking, whose definition relies on the notion of \emph{conditional
satisfaction} of c-atoms w.r.t. a pair of interpretations. 

The other two definitions, called strongly and weakly well-supported
models, are  generalizations of the notion of \emph{well-supported
models} of normal logic programs to the case of programs with c-atoms. 
As for the case of  fixpoint-based semantics, the difference between 
these two definitions is rooted in the treatment of naf atoms. 

We prove that answer sets by reduct (resp. by complement) 
are equivalent to weakly (resp. strongly) well-supported models of a program,
thus generalizing the theorem on the correspondence 
between stable models and well-supported models of a normal logic program
to the class of programs with c-atoms.  

We show that the newly defined semantics coincide with previously 
introduced semantics for logic programs with \emph{monotone} c-atoms,
and they extend the original answer set semantics of normal logic 
programs. We also study some properties of answer
sets of programs with c-atoms, and relate our definitions to several 
semantics for logic programs with aggregates presented in the
literature. 
\end{abstract}

\section{Introduction and Motivation}

Logic programming under the answer set semantics has  been 
introduced as an attractive and suitable knowledge representation language 
for AI research \cite{Baral05}, as it offers several desirable 
properties for this type of applications. Among other things, 
the language is declarative and it has a simple syntax; 
it naturally supports non-monotonic reasoning, and it
is sufficiently expressive for representing 
several classes of problems (e.g., normal logic programs
capture the class of NP-complete problems); it has solid   
theoretical foundations with a large body of building 
block results (e.g., equivalence between programs, systematic 
program development, relationships to other non-monotonic formalisms), 
which is extremely useful in the development and validation of 
large knowledge bases; 
it also has a large number of efficient computational tools. 
For further discussion of these issues, the 
interested reader is referred to the book of Baral \citeyear{Baral03},
the overview paper of Gelfond and Leone \citeyear{GelfondL02},
the paper of Marek and Truszczy\'{n}ski \citeyear{MarekT99}, 
and the paper of Niemel{\"{a}} \citeyear{Niemela99}.

A large number of extensions of logic
programming, aimed at 
improving its  usability in the context of  knowledge representation and 
reasoning, have been proposed. The {\sc Smodels} system introduces
\emph{weight} and \emph{cardinality} constraint atoms to facilitate the
encoding of constraints on atom definitions \cite{sim02}.
These constructs can be generalized to \emph{aggregates}; aggregates have
been extensively studied in the general context of logic programming 
by the work of (see, e.g., \citeR{KempS91,MumickPR90,Gelder92}),  
and further developed in recent years
(see, e.g., \citeR{Dell-ArmiFILP03,DeneckerPB01,ElkabaniPS04,FaberLP04,a-prolog,Pelov04,SonP05}). 
Both {\bf dlv}
\cite{eiter98a} and {\sc Smodels} have been 
extended to deal with 
various classes of aggregates \cite{Dell-ArmiFILP03,ElkabaniPS05}.
The semantics of these extensions have been defined either 
\emph{indirectly}, by translating programs with these extensions to normal logic
programs, or \emph{directly}, by providing new characterizations of the
concept of  answer sets for programs with such extensions. 

Each of the above mentioned extensions to logic programming has been 
introduced to facilitate the representation of a 
desirable type of knowledge in logic programming. As such, it is 
not a surprise that the focus has been on the definition of the semantics 
and little has been done to investigate the basic  
building block results for the new classes of logic programs. In this context,
the study of a uniform framework covering various classes
of extensions will provide us 
with several benefits. For example, to prove (or disprove) whether 
a basic building block result (e.g., splitting theorem) can be 
extended to the new classes of logic programs, we will need to prove 
(or disprove) this result only once; new results in the study of the generic 
framework are applicable to the study of one of the aforementioned extensions;
etc. Naturally, for these studies to be possible, a uniform framework whose 
semantical definition exhibits the behavior of various extensions of logic 
programming, needs to be developed. The main goal in this paper is to address 
this issue. 


\smallskip

The concept of logic programs with {\em abstract constraint atoms}
(or {\em c-atoms})  
has been introduced by Marek, Remmel, and Truszczy\'{n}ski 
as an elegant theoretical
framework for investigating, in a uniform fashion,
various extensions of logic programming,
including cardinality constraint atoms, weight
constraint atoms, and more general forms of  aggregates 
\cite{MarekR04,MarekT04}. Intuitively, a c-atom $A$ represents a 
\emph{constraint} on models of 
the program containing $A$---and the description of $A$ includes an
explicit description of what conditions each interpretation has to meet
in order to satisfy $A$. This view is very general, and it can be shown
to subsume the description of traditional classes of aggregates (e.g., 
\textsc{Sum, Count, Min}, etc.).\footnote{One could also argue that 
c-atoms and \emph{general} aggregates capture analogous notions.}
Thus, programs with weight constraint atoms or other  aggregates can be 
represented as logic programs with c-atoms.

The first explicit definition of answer sets for positive programs with 
arbitrary c-atoms 
(i.e., programs without the negation-as-failure operator)---called 
\emph{programs with 
set constraints} or \emph{SC-programs}---has been introduced
in the work of Marek and Remmel \citeyear{MarekR04}. 
In this work answer sets for programs with c-atoms
are defined by extending the notion of answer
sets for programs with weight constraint atoms proposed 
in the work of Niemel{\"{a}}, Simons, and Soininen \citeyear{nie99b}. 
Nevertheless, this approach provides, in certain
cases, unintuitive answer sets (see, e.g., Examples~\ref{ex0} and \ref{ex19}). In particular,
the approach of Marek and Remmel does not naturally capture some of
the well-agreed semantics for aggregates. One of our main goals 
in this paper is to investigate alternative
solutions to the problem of characterizing answer sets for programs
with arbitrary c-atoms. Our aim is to match the semantics
provided in the more recent literature for monotone c-atoms, and to avoid the 
pitfalls of the approach developed in the work of Marek and Remmel \citeyear{MarekR04}.

The concept of answer sets for programs with c-atoms has been later revisited 
by Marek and Truszczy\'{n}ski \citeyear{MarekT04}, focusing on answer sets for 
programs with \emph{monotone} 
constraint atoms,
where a c-atom $A$ is monotone if, for each pair of interpretations 
$I$ and $I'$ with $I \subseteq I'$, we have that
$I$ satisfies $A$ implies that $I'$ 
satisfies $A$. This proposal has been further extended to
the case of disjunctive logic programs with monotone
c-atoms \cite{PelovT04}. 
In another paper \cite{LiuT05}, it is extended to deal with 
convex c-atoms where a c-atom $A$ is convex if, for every pair 
of interpretations $I$ and $J$ with $I \subseteq J$, we have that 
$I$ and $J$ satisfy $A$ implies that $I'$ satisfies $A$ for every 
$I \subseteq I' \subseteq J$. This paper also proves several 
properties of programs with monotone and convex c-atoms.  
It is shown that many well-known properties of standard logic
programming under answer set semantics are preserved in the case of programs 
with monotone c-atoms. 

The main advantage of focusing  on \emph{monotone} c-atoms lies in that 
monotonicity provides a relatively simpler way for defining answer sets of logic programs with 
c-atoms. On the other hand, this restriction 
does not allow several important classes of problems  to be 
\emph{directly} expressed.
For example\footnote{
   Although variables appear in the definition of aggregates,
   they are locally quantified. As such, an aggregate literal 
   is nothing but a shorthand of a collection of ground terms. 
}, the aggregate atom $\textsc{Min}(\{X \mid p(X)\}) > 2$ 
cannot be viewed as  a monotone aggregate atom---since monotonic
extensions of the definition of $p$ might make the aggregate false;
e.g., the aggregate is true if $\{p(3)\}$ is the definition of $p$, but
it becomes false if we consider a definition containing $\{p(3),p(1)\}$.
Similarly, the cardinality constraint atom $1 \: \{a,b\} \: 1$ 
is not a monotone constraint. Neither of these two examples
can be directly encoded using monotone c-atoms. 

\smallskip
The studies in \citeA{MarekR04,MarekT04} and in \citeA{LiuT05} lead to the following 
question: ``{\em what are the alternatives to the approach to defining 
answer sets of programs with arbitrary c-atoms developed by 
\citeA{MarekR04}?}''
Furthermore, will these alternatives---if any--- capture the 
semantics of programs with monotone c-atoms proposed by \citeA{MarekT04} and avoid 
the pitfalls of the notion of answer sets for arbitrary c-atoms 
in \citeA{MarekR04}?

We present two {\em equivalent} approaches for defining 
answer sets for logic programs with arbitrary c-atoms. 
\begin{itemize}
\item The first approach is inspired by the 
 notion of \emph{conditional satisfaction}---originally
developed in \citeA{SonP05}---to characterize the semantics
of logic programs with aggregates. We generalize this
notion to the case of programs
with c-atoms. The generalization turns out to be significantly
more intuitive and easier to understand than the original
definition in \citeA{SonP05}. Using this 
notion, we define an immediate consequence operator $T_P$
for answer set checking. 
\item The second approach is inspired by the notion of \emph{well-supportedness},
proposed by Fages~\citeyear{Fages94} for normal logic programs. 
\end{itemize}
The approaches are very intuitive, and, we believe, they
improve over  the only other semantics proposed for logic programs
with arbitrary c-atoms in \citeA{MarekR04}. 

We show that  
the newly defined semantics coincide with the previously 
introduced semantics in \citeA{MarekT04} in the case of
programs with monotone c-atoms,
and they extend the original stable model semantics for normal logic 
programs. We discuss different approaches for treating 
negation-as-failure c-atoms. 
We also relate our definitions to several 
semantics for logic programs with aggregates, since the notion of
c-atom can be used to encode arbitrary aggregates. These results
show that the proposed framework naturally subsumes many existing
treatments of aggregates in logic programming.

\smallskip
\noindent
In summary, the main contributions of the paper are:
\begin{list}{$\bullet$}{\topsep=2pt \itemsep=2pt \parsep=0pt}
\item A new notion of fixpoint answer set for 
programs with arbitrary c-atoms, which is inspired by the fixpoint 
construction proposed in \citeA{SonP05} (but simpler) and 
which differs significantly from the only proposal for programs 
with arbitrary c-atoms in \citeA{MarekR04}; this will lead to
two different definitions of answer sets (answer set by reduct and answer set by 
complement); 

\item A generalization of the notion of well-supported models in \citeA{Fages94} 
to programs with arbitrary c-atoms, which---to the best of our 
knowledge---has not been investigated by any other researchers, which leads 
to the notions of weakly and strongly well-supported models; 

\item A result showing that the set of 
answer sets by reduct (resp. by complement)
is equivalent to the set of weakly (resp. strongly) well-supported models;
and

\item A number of results showing that the newly defined notions of answer sets 
capture the answer set semantics of various extensions to logic 
programming, in those cases all the previously proposed semantics agree. 
\end{list}

\smallskip
The rest of this paper is organized as follows. Section~\ref{prel} presents preliminary 
definitions, including the syntax of the language of logic programming with
c-atoms, the basic notion of satisfaction, and the notion of answer set 
for programs with monotone c-atoms in \citeA{MarekT04} and 
for positive programs with arbitrary c-atoms in \citeA{MarekR04}. 
Section~\ref{firstapp} presents our first
approach to defining answer sets for logic programs with arbitrary c-atoms
based on a fixpoint operator, while
Section~\ref{secondapp} introduces an alternative definition based on 
well-supportedness. Section~\ref{general} extends the semantics to 
programs with arbitrary c-atoms in the head of rules. 
Section~\ref{discuss} relates the semantics presented in this paper with 
early work on abstract constraint atoms and aggregates. 
Section~\ref{conc} provides conclusions and 
future work. Proofs of theorems and propositions are deferred to the appendix. 

\section{Preliminaries---Logic Programs with Abstract Constraint Atoms}\label{prel}

We follow the syntax used in \citeA{LiuT05} to define 
programs with abstract constraint atoms.
Throughout the paper, we assume a fixed propositional language 
$\cal L$ with a countable set ${\cal A}$ of 
propositional atoms. 

\subsection{Syntax} \label{syntax}
An \emph{abstract constraint atom} (or {\em c-atom})
is an expression of the form $(D,C)$, where $D \subseteq {\cal A}$ 
is a set of atoms (the \emph{domain} of the c-atom), 
and $C$ is a collection of sets of atoms belonging to $D$,
i.e., $C \subseteq 2^{D}$
(the \emph{solutions} of the c-atom). 
Intuitively, a c-atom $(D,C)$ is  a constraint 
on the set of atoms $D$, and $C$ represents its admissible solutions. 
 Given 
a c-atom $A = (D,C)$, we use $A_d$ and $A_c$ to denote
$D$ and $C$, respectively. 

A c-atom of the form $(\{p\},\{\{p\}\})$ is called 
an {\em elementary} c-atom and will be simply written as $p$.  
A c-atom of the form $({\cal A},\emptyset)$, representing a
constraint which does not admit any solutions, 
 will be denoted by $\bot$.
A c-atom $A$ is said to be {\em monotone} 
if for every $X \subseteq Y \subseteq A_d$, $X \in A_c$ implies 
that $Y \in A_c$.

A {\em rule} is of the form 
\begin{equation}\label{rule}
A \leftarrow A_1,\ldots,A_k,\naf A_{k+1},\ldots,\naf A_n
\end{equation}
where $A$, $A_j$'s are c-atoms. The literals $\naf A_j$ ($k < j \le n$)
are called \emph{negation-as-failure c-atoms} (or \emph{naf-atoms}). 
For a rule $r$ of the form (\ref{rule}), we define:
\begin{itemize}
\item $head(r) = A$, 
\item $pos(r) = \{A_1, \dots, A_k\}$, 
\item $neg(r) = \{A_{k+1}, \dots, A_n\}$,
\item $body(r) = \{A_1,\ldots,A_k,\naf A_{k+1},\ldots,\naf A_n\}$.
\end{itemize}
For a program $P$, $hset(P)$ denotes the set 
$\cup_{r \in P} head(r)_d$.

We recognize special types of rules:
\begin{enumerate}
\item A rule $r$ is {\em positive} if $neg(r) = \emptyset$;
\item A rule $r$ is  {\em basic} if $head(r)$ is an elementary c-atom;
\item A rule $r$ is a {\em constraint} rule if $head(r) = \bot$.
\end{enumerate}
A \emph{logic program with c-atoms} 
(or \emph{logic program}, for simplicity)\footnote{Whenever we
want to refer to traditional logic programs (without c-atoms), we will
explicitly talk about \emph{normal logic programs}.} 
is a set of rules. A program $P$ is called a \emph{basic} program 
if each rule $r \in P$ is a basic or a constraint rule. $P$ is said to be 
\emph{positive} if every rule in $P$ is positive. 
$P$  is {\em monotone} (resp. {\em naf-monotone}) if each c-atom 
occurring in $P$ (resp. in a naf-atom in $P$) is monotone. 
Clearly, a monotone program is also naf-monotone.

\subsection{Models and Satisfaction}

In this subsection, we introduce the basic definitions for the study 
of logic programs with constraints. We will begin with the definition 
of the satisfaction of c-atoms. We then introduce the 
notion of a model of programs with c-atoms. 

\subsubsection{Satisfaction of C-Atoms}
A set of atoms $S \subseteq {\cal A}$ satisfies a c-atom $A$,
denoted by $S \models A$, if 
$A_d \cap S \in A_c$. $S$ satisfies  $\naf A$,
denoted by $S \models \naf A$, if 
$A_d \cap S \not\in A_c$.

It has been shown in \citeA{MarekR04} and in \citeA{MarekT04} that 
the notion of c-atom is more general than extended
atoms such as cardinality constraint atoms and aggregate atoms; thus,
c-atoms  can be used to conveniently represent 
weight constraints, cardinality constraints \cite{sim02}, 
and various other classes  of aggregates, such 
as maximal cardinality constraints.  
For example, 
\begin{itemize}
\item Let us consider an arbitrary choice atom of the
form $L \{ p_1, \dots, p_k, not\:q_1, \dots, not\:q_h\} U$; this
can be represented by the c-atom $(A,S)$ where:
	\begin{itemize}
	\item $A = \{p_1, \dots, p_k, q_1, \dots, q_h\}$
	\item $S= \left\{\:\: T \subseteq A \:\:\mid\:\:
					L \leq | (T \cap \{p_1,\dots,p_k\}) 
\cup (\{q_1,\dots,q_h\}\setminus T) | \leq U
				      				\:\:\right\}$
	\end{itemize}

\item Let us consider an arbitrary aggregate of the form
	$F\{ v \:|\: p(v)\} \oplus V$ where $F$ is a
	set function (e.g., {\sc Sum}, {\sc Avg}),
	$V$ is a number, and $\oplus$ is a comparison operation (e.g., $\ge, >, \ne$). 
	This can be represented by the
	c-atom $(A,S)$, where:
	\begin{itemize}
	\item $A = \{p(a) \:|\: p(a) \in {\cal A}\}$
	\item $S = \{T  \:|\: T \subseteq A, F(T) \oplus V\}$
	\end{itemize}

\end{itemize}

\begin{example}
{\rm
Let us consider the aggregate
 $\textsc{sum}(\{X \mid p(X)\}) {\ge} -1$, defined in a language where
${\cal A} = \{p(1), p(-2)\}$. From the considerations above, we have that
this aggregate can be represented by the c-atom $(\{p_1),p(-2)\}, S)$ where
\[ S =  \{T  \:|\: T \subseteq \{p(1),p(-2)\}, \textsc{sum}(T) \geq -1\} = \{ \emptyset, \{p(1)\}, \{p(-2),p(1)\}\}\]
}
\hfill $\Box$
\end{example}

\begin{example}
{\rm
Let us consider the cardinality constraint atom 
$1 \: \{p(1),p(-1)\} \: 1$. This can be represented by the
c-atom $(\{p(1),p(-1)\}, S)$ where 
\[ S = \left\{\:\: T  \:\:\mid\:\: T \subseteq \{p(1),p(-1)\},
					1 \leq | (T \cap \{p(1),p(-1)\} | \leq 1
				      				\:\:\right\} = \{ \{p(1)\}, \{p(-1)\} \}\]
}
\hfill $\Box$
\end{example}
C-atoms allow us to compactly represent properties
that would require complex propositional combinations of traditional 
aggregates. E.g., a condition like \emph{``either all elements or no elements
of the set $\{a,b,c,d\}$ are true''} can be simply written as the single
c-atom  $(\{a,b,c,d\}, \{\emptyset,\{a,b,c,d\}\})$. Further motivations 
behind the use of c-atoms can be found in \citeA{MarekR04} and \citeA{MarekT04}.

In the rest of the paper, we will often use in our
examples the notation of cardinality 
constraint atoms, weight constraint atoms, or general aggregate atoms 
 instead of c-atoms,
whenever no confusion is possible.

\subsubsection{Models}

A set of atoms $S$ satisfies the body of a rule $r$ of
the form (\ref{rule}), denoted by $S \models body(r)$,
 if $S \models A_i$ for $i=1,\ldots,k$ 
and $S \models \naf A_j$ for $j=k+1,\ldots,n$. $S$ satisfies a
rule $r$ if it satisfies $head(r)$ or if  it does not
satisfy $body(r)$.

A set of atoms $S$ is a {\em model} of a program $P$ 
if $S$ satisfies every rule of $P$.
$M$ is a \emph{minimal} model of $P$ if it is a model of $P$ 
and there is no proper subset of $M$ which is 
also a model of $P$. 
In particular, programs
may have more than one minimal model (see, for example,
 Example~\ref{ex4}).

Given a program $P$, a set of atoms $S$ is said to support 
an atom $a \in {\cal A}$ if there exists some rule $r$ in $P$ 
and $X \in head(r)_c$ such that the following conditions are met:
\begin{itemize}
\item $S \models body(r)$, 
\item $X \subseteq S$, and 
\item $a \in X$.
\end{itemize}

\begin{example} 
\label{ex2}
{\rm
Let $P_1$\footnote{Remember that the notation $p$ is a short form for
	the c-atom $(\{p\},\{\{p\}\})$.} be the program 
\[
\begin{array}{lllllcll}
  p(a) & \leftarrow  &  \\
 p(b)& \leftarrow &  \\
 p(c)&   \leftarrow &  q\\
  q & \leftarrow & {\textnormal{\sc Count}(\{X \mid p(X)\}) > 2} 
\end{array}
\]
The aggregate notation  $\textnormal{\sc Count}(\{X {\mid} p(X)\}) > 2$
represents the c-atom $(D,\{D\})$ 
where $D=\{p(a), p(b), p(c)\}$. 
$P_1$ has two models:
\[\begin{array}{lcr}
M_1 = \{p(a), p(b), p(c), q\} & \hspace{1cm} &
M_2 = \{p(a), p(b)\}
  \end{array}
\]
$M_2$ is a minimal model of $P_1$, while $M_1$ is not.
\hfill $\Box$
}
\end{example}

\begin{example}
\label{ex3}
{\rm
Let $P_2$ be the program 
\[
\begin{array}{lcl}
p(1) & \leftarrow & \\
p(-1)&   \leftarrow&  p(2) \\
p(2) & \leftarrow & \textsc{Sum}(\{X \mid p(X)\}) \ge 1  
\end{array}
\]
The aggregate notation
$\textsc{Sum}(\{X {\mid} p(X)\}) {\ge} 1$ represents the c-atom 
$(D,C)$ where 
\[\begin{array}{lclcl}
D & = &  \{p(1),p(2),p(-1)\} & \hspace{.5cm} & \textit{and} \\
C & = & \multicolumn{3}{l}{\{\{p(1)\}, \{p(2)\}, \{p(1), p(2)\}, \{p(2),p(-1)\}, \{p(1),p(2),p(-1)\}\}}
\end{array}
 \]
Because of the first rule, 
any model of $P_2$ will need to contain $\{p(1)\}$.
It is easy to see that $\{p(1),p(-1)\}$ and 
$\{p(1),p(2),p(-1)\}$ are models of $P_2$ but 
$\{p(1),p(2)\}$ is not a model of $P_2$. 
\hfill $\Box$
}
\end{example}
\begin{example}
\label{ex4}
{\rm
Let $P_3$ be the program 
\[
\begin{array}{lcl}
p & \leftarrow & (\{q\}, \{\emptyset\})\\
q & \leftarrow & (\{p\}, \{\emptyset\})
\end{array}
\]
$P_3$ has three models $\{p\}$, $\{q\}$, and $\{p,q\}$, of which 
$\{p\}$ and $\{q\}$ are minimal.
\hfill$\Box$
}
\end{example}

\subsection{Previously Proposed Semantics}
In this section, we will overview the semantical characterizations for
programs with c-atoms proposed in the existing literature. In particular,
we will review the notion of answer sets for monotone programs (i.e., 
program that contain only  monotone c-atoms), as defined in \citeA{MarekT04}.
A formal 
comparison between these semantics and the novel approach we propose
in this paper is described in Section~\ref{discuss}.


Given a set of atoms $S$, 
a rule $r$ is {\em applicable} in $S$ if $S \models body(r)$. 
The set of applicable rules in $S$ is denoted by $P(S)$. 
A set $S'$ is {\em nondeterministically one-step provable}
from $S$ by means of $P$ if $S' \subseteq hset(P(S))$ and 
$S' \models head(r)$ for every $r \in P(S)$. 
The {\em nondeterministic one-step provability operator} $T^{nd}_P$ 
is a function from $2^{\cal A}$ to $2^{2^{\cal A}}$ such that 
for every $S \subseteq {\cal A}$, $T^{nd}_P(S)$ consists of all sets $S'$
that are nondeterministically one-step provable from $S$ by means of $P$. 

A \emph{$P$-computation} is
a sequence $t=(X_n)_{n=0,1,2,\dots}$ where $X_0=\emptyset$ and for every non-negative 
integer $n$, 
\begin{itemize}
\item [] ({\em i}) $X_n \subseteq X_{n+1}$, and 
\item [] ({\em ii}) $X_{n+1} \in T^{nd}_P(X_n)$
\end{itemize}
$S_t = \cup_{n=0}^\infty X_i$ is called the {\em result} of the computation $t$. 
A set of atoms $S$ is a {\em derivable model} of $P$ if there exists 
a $P$-{\em computation} $t$ such that $S = S_t$. 
The Gelfond-Lifschitz reduct for normal logic programs is generalized to 
monotone programs as follows. 

\begin{definition}
Let $P$ be a monotone program. For a set of atoms $M$, the {\em reduct} of 
$P$ with respect to $M$, denoted by $P^M$, is obtained from $P$ by 
\begin{enumerate}
\item removing from $P$ every rule containing in the body a literal 
$\naf A$ such that $M \models A$; and 

\item removing all literals of the form $\naf A$ from the remaining rules.
\end{enumerate}
\end{definition}
Answer sets for monotone programs are defined next. 
\begin{definition} \label{as-marek}
A set of atoms $M $ is an {\em answer set} of a monotone program $P$ if $M$ is 
a derivable model of the reduct $P^M$. 
\end{definition}
The next example shows that, for programs with non-monotone c-atoms, 
Definition \ref{as-marek} is, in general, not applicable. 
\begin{example} \label{ex41}
{\rm Consider the program $P_3$ from Example \ref{ex4}.
We can check that this program does not allow the construction of any
$P_3$-computation. In fact,  
$T^{nd}_{P_3}(\emptyset) = \{\{p,q\}\}$ and 
$T^{nd}_{P_3}(\{p,q\}) = \{\emptyset\}$. Hence, $\{p\}$ would not be an 
answer set of $P_3$ (according to Definition \ref{as-marek}) 
since it is not a derivable model of the reduct of $P_3$ 
with respect to $\{p\}$ (which is $P_3$). 

On the other hand, it is easy to see that $P_3$ is intuitively equivalent to the 
normal logic program $\{p \leftarrow \naf q, q \leftarrow \naf p\}$. 
As such, $P_3$ should 
accept $\{p\}$ as one of its answer sets. 
\hfill $\Box$
}
\end{example}
The main reason for the inapplicability of Definition \ref{as-marek} 
lies in that the nondeterministic one-step provability operator 
$T^{nd}_P$ might become non-monotone in the presence of non-monotone c-atoms. 

\subsection{Answer Sets for Positive Programs} 

Positive programs are characterized by the lack of negation-as-failure atoms.
Positive programs with arbitrary c-atoms have been investigated in
\citeA{MarekR04}, under the name of \emph{SC-programs}.
Let us briefly 
review the notion of answer sets for SC-programs---which, in turn,  is a generalization 
of the notion of answer sets for logic programs with weight constraints, as presented  
in \citeA{nie99b}. 
A detailed comparison between the approach in  \citeA{MarekR04} 
and our work is given in Section \ref{discuss}.

For a c-atom $A$, the closure of $A$, denoted by 
$\widehat{A}$, is the c-atom
\[
(\:A_d,\:\: \{Y \mid Y \subseteq A_d, \: \exists Z. \: (Z \in A_c,  Z \subseteq Y)\}\:)
\] 
Intuitively, the closure is constructed by including all the supersets of the
existing solutions.

A c-atom $A$ is said to be \emph{closed} if $A = \widehat{A}$. 
A rule of the form (\ref{rule}) is a Horn-rule if ({\em i}) its head is an elementary 
c-atom; and ({\em ii}) each c-atom in the body is 
 a closed c-atom. 
A SC-program $P$ is said to be a Horn SC-program if each rule in $P$ 
is a Horn-rule. The one-step provability operator, defined by 
$T_P(X) = \{a \mid \exists r \in P, \: head(r) = a, \: X \models body(r)\}$, 
associated to a Horn SC-program $P$ is monotone. Hence, every 
Horn SC-program $P$ has a least fixpoint $M^P$ which is the only minimal 
model of $P$ (w.r.t. set inclusion). Given a set of atoms $M$ and a
SC-program $P$, the \emph{NSS-reduct} of $P$ with respect to $M$, denoted by 
$NSS(P,M)$, is obtained from $P$ by 
\begin{list}{}{\topsep=1pt \parsep=0pt \itemsep=1pt}
\item[({\em i})] 
removing all rules whose body is not satisfied by $M$; and 
\item[({\em ii})] replacing each rule 
\[
A \leftarrow e_1,\ldots,e_n,A_1,\ldots,A_m
\]
where $e_i$'s are elementary c-atoms and 
$A_j$'s are non-elementary c-atoms by the set of rules
\[
\{a \leftarrow 
	e_1,\ldots,e_n,\widehat{A_1},\ldots,\widehat{A_m} 
  \mid 
        a \in A_d \cap M \}
\]
\end{list}
A model $S$ of a program $P$ is an answer set of $P$ 
if it is the least fixpoint of the one-step provability 
operator of the program $NSS(P,S)$, i.e., $S = M^{NSS(P,S)}$.
It sometimes yields 
answer sets that are not accepted by other extensions to 
logic programming. The next example illustrates this
point.

\begin{example}
\label{ex0}
{\rm
Consider the program $P_4$:
\[
\begin{array}{lll} 
c & \leftarrow & \\
a & \leftarrow & (\{a,c\}, \{\emptyset, \{a,c\}\}) 
\end{array}
\]
We have that $M_1 = \{c\}$ and $M_2 = \{a,c\}$ are models of $P_4$.  
Furthermore, $NSS(P_4, M_1)$ is the program 
\[
\begin{array}{lll} 
c & \leftarrow & 
\end{array}
\]
and $NSS(P_4, M_2)$ consists of the rules 
\[
\begin{array}{lll} 
c & \leftarrow & \\
a & \leftarrow & (\{a,c\}, \{\emptyset, \{a\}, \{c\}, \{a,c\})\} \\
\end{array}
\] 
It is easy to see that $M_1 = M^{NSS(P,M_1)}$ and $M_2 = M^{NSS(P,M_2)}$. 
Thus, observe that $P_4$ has a non-minimal answer set $\{a,c\}$ 
according to \citeA{MarekR04}. Note that the program $P_4$ can be viewed as 
the following program with aggregates 
\[
\begin{array}{lll} 
c & \leftarrow & \\
a & \leftarrow & \textsc{Count}(\{a, c\}) \ne 1 
\end{array}
\]
which does not have $\{a,c\}$ as an answer set under most of the proposed 
semantics for aggregates \cite{DeneckerPB01,FaberLP04,Ferraris05a,Pelov04,SonP05}.
Furthermore, all these approaches accept $\{c\}$ as the only answer 
set of this program.}
\hfill$\Box$
\end{example}

\section{Answer Sets for Basic Programs: A Fixpoint Based Approach} \label{firstapp}

In this section, we define the notion of answer sets of \emph{basic} programs. 
In this approach, we follow the traditional way for defining answer sets
of logic programs, i.e., by:
\begin{enumerate}
\item  first characterizing the semantics of  positive programs 
(Definition \ref{basic-pos-a.s.}), and then 
\item  extending it to deal with naf-atoms 
(Definitions \ref{def-basic} and \ref{reduct-a.s.}). 
\end{enumerate}

\subsection{Answer Sets for Basic Positive Programs} 

Example~\ref{ex4} shows that a basic positive program 
might have more than one minimal model. This leads us 
to define a $T_P$-like operator for answer set checking,
whose construction  is based on the 
following observation.  

\begin{quote}
\begin{observation} \label{observ1}
   Let $P$ be a propositional normal logic program (i.e., 
    without c-atoms)\footnote{For a rule $r$ from a normal logic program $P$,\\
     \centerline{$ a \leftarrow a_1,\ldots,a_n,\naf b_1,\ldots,\naf b_m$}\\
   $head(r)$, $pos(r)$, and $neg(r)$ denote $a$, 
    $\{a_1,\ldots,a_n\}$, and $\{b_1,\ldots,b_m\}$, respectively. 
} and let $R, S$ be two sets 
of atoms. Given a set of atoms $M$, we define the operator $T_P(R,S)$ and the 
monotone sequence of interpretations
 $\langle I_i^M \rangle_{i=0}^\omega$ as follows.
\[
T_P(R,S) = \left \{  a
\begin{array}{l|ll}
 & & \exists r \in P \: :  \: head(r) = a, \\
 & & pos(r) \subseteq R, neg(r) \cap S = \emptyset\\
\end{array}
\right \}
\]
\[\begin{array}{lcr}
I_0^M = \emptyset & \hspace{1cm} & I_{i+1}^M = T_P(I_i^M, M) \hspace{1cm} (i \ge 0)
  \end{array}
\]
Let us denote with $I_\omega^M$ the limit of this sequence of interpretations.
It is possible to prove that $M$ is an answer set of $P$ w.r.t.~\citeA{GelfondL88} iff
$M = I_\omega^M$\/.
\end{observation}
\end{quote}

As we can see from the above observation, the (modified) 
consequence operator $T_P$ takes two sets of atoms, $R$ and $S$, 
as its arguments, and generates one set of atoms which could be viewed 
as the consequences of $P$ given that $R$ is true and $S$ is assumed 
to be an answer set of $P$. It is easy to see that $T_P$ is monotone
w.r.t. its first argument, i.e., if $R \subseteq V$, then
$T_P(R,S) \subseteq T_P(V,S)$. Thus, the sequence 
$\langle I_j^M \rangle_{j=0}^\omega$ 
is monotone and converges to  $I_\omega^M$ for a given $S$. 
We will next show how  $T_P$ can be generalized to programs with c-atoms. 

\smallskip
Observe that the definition of $T_P$ requires that 
$pos(r) \subseteq R$ or, equivalently, $R \models pos(r)$. For normal 
logic programs, this is sufficient to guarantee the monotonicity of 
$T_P(\cdot,S)$. If this definition is naively generalized to the case of
programs with c-atoms, the monotonicity 
of $T_P(.,S)$ is guaranteed only under certain circumstances,
e.g.,  when c-atoms in $pos(r)$ are monotone. 
To deal with arbitrary c-atoms, we need to  introduce the notion of 
\emph{conditional satisfaction} of a c-atom.

\begin{definition}
[Conditional Satisfaction] 
\label{cond-sat}
Let $M$ and $S$ be sets of atoms. The set $S$ 
{\em conditionally satisfies} a c-atom $A$ w.r.t. $M$,
denoted by $S \models_M A$, if 
\begin{enumerate}
\item $S\models A$ and, 
\item for every $I\subseteq A_d$
such that $S \cap A_d \subseteq I$ and $I \subseteq M \cap A_d$, 
we have that $I \in A_c$.
\end{enumerate}
\end{definition}
Observe that this notion of conditional satisfaction has
been inspired by the conditional satisfaction used to
characterize aggregates in~\citeA{SonP05}, 
but it is significantly simpler. 

We say that $S$ conditionally satisfies a set 
of c-atoms $V$ w.r.t. $M$, denoted by $S \models_M V$,
if $S \models_M A$ for every $A \in V$. 
Intuitively, $S \models_M V$ implies that $S' \models V$ 
for every $S'$ such that $S \subseteq S' \subseteq M$.
Thus, conditional satisfaction ensures that if the body of a rule 
is satisfied in $S$ then it is also satisfied in $S'$, 
provided that $S \subseteq S' \subseteq M$.
This allows us to generalize the operator $T_P$ defined in Observation 1
as follows. For a set of atoms $S$ and a positive
basic program $P$, let
\[
T_P(S, M) = \left\{a \:\: \begin{array}{|ll}
      &  \exists r\in P\: : \: S \models_M pos(r),\:\:
          head(r) = (\{a\}, \{\{a\}\})\end{array}\right\}
\]
The following proposition holds. 
\begin{proposition}
\label{prop-tp}
Let $M$ be a model of $P$, and let 
$S \subseteq U \subseteq M$. Then
$T_P(S, M) \subseteq T_P(U,M) \subseteq M.$
\end{proposition}

\noindent 
Let $T_P^0(\emptyset,M) = \emptyset$
and, for $i \ge 0$, let
$$T^{i+1}_P(\emptyset,M) = T_P(T^i_P(\emptyset,M), M)$$
Then, the following corollary is a natural consequence
of Proposition~\ref{prop-tp}.
\begin{corollary}
\label{cr-tp}
Let $P$ be a positive, basic program and $M$ be a model of $P$.
Then, we have
$$T_P^0(\emptyset,M) \subseteq T_P^{1}(\emptyset,M) \subseteq  
\dots \subseteq M$$
\end{corollary}
The above corollary implies that the sequence 
$\langle T^i_P(\emptyset,M) \rangle_{i=0}^\infty$ 
is monotone and limited (w.r.t. set inclusion) by $M$. Therefore, it
converges to a fixpoint. We denote this fixpoint by
$T^\infty_P(\emptyset, M)$.

\begin{definition} \label{basic-pos-a.s.}
Let $M$ be a model of a basic positive program $P$. 
$M$ is an \emph{answer set} of $P$ iff $M = T^\infty_P(\emptyset, M)$.
\end{definition}
Observe that the constraint rules present in $P$ (i.e., rules whose
head is $\bot$) do  not contribute to the 
construction performed by $T_P$; nevertheless, the requirement
that $M$ should be a model of $P$ implies that all the constraint
rules
will have to be satisfied by each answer set.
We illustrate Definition \ref{basic-pos-a.s.} in the next examples.
\begin{example}
{\rm
Consider the program  $P_1$ from Example \ref{ex2}. 
\begin{itemize}
\item $M_1 = \{p(a),p(b)\}$ is an answer set of $P_1$ 
since:
\[
\begin{array}{lll}
T^0_{P_1}(\emptyset,M_1) & = &  \emptyset \\
T^{1}_{P_1}(\emptyset,M_1) & = & \{p(a), p(b)\} = M_1 \\ 
T^{2}_{P_1}(\emptyset,M_1) & = & T_{P_1}(\{p(a), p(b)\}, M_1) = M_1 \\ 
\end{array}
\]
\item $M_2 = \{p(a),p(b),p(c), q\}$ is not an answer set of $P_1$,
since:
\[
\begin{array}{lll}
T^0_{P_1}(\emptyset,M_2) & = &  \emptyset \\
T^{1}_{P_1}(\emptyset,M_2) & = & \{p(a), p(b)\} = M_1\\ 
T^{2}_{P_1}(\emptyset,M_2) & = & T_{P_1}(\{p(a), p(b)\}, M_2) = M_1 \\ 
\end{array}
\]
\hfill$\Box$
\end{itemize}
}
\end{example}

\begin{example}
{\rm
Consider again the program $P_3$ (Example \ref{ex4}). 
Let $M_1 = \{p\}$ and $M_2 = \{q\}$. We have that 
\[
\begin{array}{llllllll}
T^0_{P_3}(\emptyset,M_1) & = &  \emptyset & & 
	T^0_{P_3}(\emptyset,M_2) & = &  \emptyset \\
T^{1}_{P_3}(\emptyset,M_1) & = & \{p\} = M_1 && 
	T^{1}_{P_3}(\emptyset,M_2) & = & \{q\} = M_2\\
\end{array}
\]
Thus, both $\{p\}$ and $\{q\}$ are answer sets of $P_3$. 
On the other hand, for $M = \{p,q\}$, we have that 
$T^{1}_{P_3}(\emptyset,\{p,q\}) = \emptyset$
because $\emptyset \not\models_M (\{q\},\{\emptyset\})$ 
and $\emptyset \not\models_M (\{p\},\{\emptyset\})$.
Hence, $\{p,q\}$ is not an answer set of $P_3$.
\hfill$\Box$
}
\end{example}
We conclude this section by observing that the answer
sets obtained from the above construction are minimal
models.

\begin{corollary}
\label{cr-minm} 
Let $P$ be a positive basic program and
$M$ be an answer set of $P$. Then, $M$ is a minimal 
model of $P$.
\end{corollary}
The next example shows that not every positive program has an answer set. 
\begin{example}
{\rm
Consider $P_2$ (Example \ref{ex3}). Since 
answer sets of positive programs are \emph{minimal} models
(Corollary~\ref{cr-minm}) and $M = \{p(1),p(-1)\}$ 
is the only minimal model of $P_2$, we have that $M$ is 
the only possible answer set of $P_2$. Since
\[
\begin{array}{lll}
T^0_{P_2}(\emptyset,M) & = &  \emptyset \\
T^{1}_{P_2}(\emptyset,M) & = & \{p(1)\} \\ 
T^{2}_{P_2}(\emptyset,M) & = & T_{P_2}(\{p(1)\}, M) = \{p(1)\} \\ 
\end{array}
\]
we can conclude that $M$ is not an answer set 
of $P_2$. Thus, $P_2$ does not have answer sets. 
\hfill $\Box$
}
\end{example}
The example highlights that supportedness, in our approach, is not
a sufficient condition for being an answer set---$M'=\{p(1),p(-1),p(2)\}$ is
a supported model, but it is not accepted as an answer set. The reason for rejecting
$M'$ is the fact that the element $p(2)$ is essentially self-supporting itself
(cyclically) in $M'$. Note that $M'$ is rejected, as an answer set, in most 
approaches to aggregates in logic programming---e.g., the approach in~\citeA{FaberLP04}
rejects $M'$ for not being a minimal model of the FLP-reduct of the program.

\subsection{Answer Sets for Basic Programs} 

We will now define answer sets for basic programs, i.e., programs with 
elementary c-atoms in the head of the rules, and rule bodies composed
of c-atoms and  naf-atoms.

In the literature, two main approaches have been
considered to deal with negation of aggregates and of
other complex atoms. Various  extensions of logic 
programming (e.g., weight constraints in \citeA{sim02} and aggregates in \citeA{FaberLP04}) support
negation-as-failure atoms by replacing each naf-atom $\naf A$ with a c-atom 
$A'$, where $A'$ is obtained from $A$ by replacing the predicate relation 
of $A$ with its ``negation''. For example, following this approach,
the negated cardinality constraint atom
$\naf 1 \:\{a,b\}\: 1 $ can be replaced by 
$(\{a,b\}, \{\emptyset, \{a,b\}\})$. Similarly, the 
negated aggregate atom 
$\naf \textnormal{\sc Sum}(\{X \mid p(X)\}) \ne 5$ can be replaced 
by $\textnormal{\sc Sum}(\{X \mid p(X)\}) = 5$. 

On the other hand, other researchers (see, e.g., \citeR{MarekT04,Ferraris05a}) have
suggested to handle 
naf-atoms  by using a form of \emph{program reduct}---in 
the same spirit as in \citeA{GelfondL88}. 

\smallskip

Following these perspectives, we study two different 
approaches for dealing with naf-atoms, described in the next two
subsections. It is worth mentioning that 
both approaches coincide in the case of  monotone programs 
(Proposition~\ref{prop-equiv}). 
\subsubsection{Negation-as-Failure by Complement} 

In this approach, we treat a naf-atom $\naf A$ by replacing it
with its {\em complement}. We define 
the notion of  \emph{complement} of a c-atom as follows. 
\begin{definition}\label{compldef}
The complement $\bar{A}$ of a c-atom $A$ is the 
c-atom $(A_d,2^{A_d} \setminus A_c)$.
\end{definition}
We next define the \emph{complement} of a program $P$. 
\begin{definition}
Given a basic program $P$, we define ${\cal C}(P)$
(the complement of $P$) 
to be the program obtained from $P$ by replacing each
occurrence of $\naf A$ in $P$ with the complement of $A$.
\end{definition}
The program  ${\cal C}(P)$ is a basic positive program, whose answer sets
have been defined in Definition \ref{basic-pos-a.s.}.
This allows us to
define the notion of  answer sets of basic programs as follows. 
\begin{definition} \label{def-basic}
A set of atoms $M$ is an {\em answer set by complement} 
of a basic program $P$ iff it is an answer 
set of ${\cal C}(P)$.
\end{definition}
It is easy to see that each answer set of a program $P$ is indeed
a  minimal model of $P$.
\begin{example}\label{pippo1}
{\rm
Let us consider the program $P_5$, which consists of the following rules:
\[
\begin{array}{lll}
a & \leftarrow & \\
c & \leftarrow & \naf (\{a,b\},\{\{a,b\}\}) \\
\end{array}
\]
The complement of $P_5$ is 
\[
\begin{array}{lll}
a & \leftarrow & \\
c & \leftarrow & (\{a,b\},\{\emptyset, \{a\}, \{b\}\}) \\
\end{array}
\]
which has $\{a,c\}$ as its only answer set. Thus, $\{a,c\}$ 
is an answer set by complement of $P_5$.
\hfill$\Box$
}
\end{example}

\begin{example} \label{complement}
{\rm
Let $P_6$ be the program 
\[\begin{array}{lcl}
c&  \leftarrow&  \naf 1 \{a,b\} 1\\
a&  \leftarrow&  c\\
b&  \leftarrow&  a
  \end{array}
\] 
We have that ${\cal C}(P_6)$ is the program
\[\begin{array}{lcl}
c&  \leftarrow&  (\{a,b\}, \{\emptyset, \{a,b\}\})\\
a&  \leftarrow&  c\\
b&  \leftarrow&  a
  \end{array}
\]
This program does not have an answer set (w.r.t. Definition \ref{basic-pos-a.s.});
thus $P_6$ does not have an answer set by complement.
\hfill $\Box$
}
\end{example}

\subsubsection{Negation-as-Failure by Reduction}

Another approach for dealing with naf-atoms is to adapt the 
Gelfond-Lifschitz reduction of normal logic programs \cite{GelfondL88} 
to programs with c-atoms---this approach has been considered
in \citeA{MarekT04} and \citeA{Ferraris05a}. We 
can generalize this approach to programs with arbitrary c-atoms as follows. For 
a basic program $P$ and a set of atoms $M$, 
the {\em reduct of P w.r.t. M} 
($P^M$)  is the set of rules obtained by 
\begin{enumerate}
\item removing all rules containing $\naf A$ s.t. 
$M \models A$; and 
\item removing all $\naf A$ from the remaining rules.
\end{enumerate}
The program $P^M$ is a positive basic  program. Thus,
we can define answer sets for $P$ as follows:
\begin{definition} \label{reduct-a.s.}
A set of atoms $M$ is an {\em answer set by reduct} of $P$ iff 
$M$ is an answer set of $P^M$ (w.r.t. Definition \ref{basic-pos-a.s.}).
\end{definition}
%
\begin{example}
\label{ex41b}
{\rm
Let us reconsider the program $P_5$ from Example~\ref{pippo1} and let us
consider $M=\{a,c\}$. If we perform a reduct, we are left with the rules
\[
\begin{array}{lll}
a & \leftarrow & \\
c & \leftarrow &
\end{array}
\]
whose minimal model is $M$ itself. Thus, $M$ is an answer set 
by reduct of the program $P_5$.
\hfill $\Box$
}
\end{example}
The next example shows that this approach can lead to different answer
sets than the case of negation by complement (for non-monotone programs).
\begin{example} 
\label{ex5}
{\rm 
Consider the program $P_6$ from Example \ref{complement}.
Let $M = \{a,b,c\}$. The reduct of $P_6$ w.r.t. $M$ is the program 
\[\begin{array}{lcl}
c &\leftarrow\\
a&  \leftarrow&  c\\
b&  \leftarrow&  a
  \end{array}
\] 
which has $M$ as its answer set, i.e., 
$M$ is an answer set by reduct of $P_6$.
\hfill $\Box$
}
\end{example}
One consequence of the negation by reduct  approach is the
fact that it might lead to \emph{non-minimal} 
answer sets---in the presence of non-monotone atoms. For instance, 
if we replace the atom $\textnormal{\sc Count}(\{X \mid p(X)\})>2$ 
in $P_1$ with $\naf \textnormal{\sc Count}(\{X \mid p(X)\})\le 2$,
the new program is (by replacing the aggregate with a c-atom):
\[\begin{array}{lcl}
p(a) & \leftarrow\\
p(b) & \leftarrow \\
p(c) & \leftarrow & q\\
q & \leftarrow & not\:\: \left(\{p(a),p(b),p(c)\}, \: \left\{\begin{array}{l}
							\emptyset, \{p(a)\}, \{p(b)\},	\{p(c)\},\\
							 \{p(a),p(b)\}, \{p(b),p(c)\}, \{p(a),p(c)\}
							     \end{array}\right\}\right)
  \end{array}
\]
This program admits the following two interpretations as
answer sets by reduct:  $M_1=\{p(a),p(b),p(c),q\}$ and
$M_2=\{p(a), p(b)\}$. Since $M_2 \subseteq M_1$, we have
that a non-minimal answer set exists.

This result indicates that, for certain  programs with 
c-atoms, there might be different ways to treat naf-atoms, leading
to different semantical characterizations. This 
problem has been mentioned in \citeA{Ferraris05a}. 
Investigating other methodologies for dealing with naf-atoms is an 
interesting topic of research, that we plan to pursue in the future.

\subsection{Properties of Answer Sets of Basic Programs}
We will now show that the notion of answer sets
for basic programs with c-atoms is a natural generalization of 
the notions of answer sets for normal logic programs. We 
prove that answer sets of basic positive programs are minimal 
and supported models and characterize situations in which these 
properties hold for basic programs. We begin with a result 
stating that, for the class of naf-monotone programs, 
the two approaches for dealing with naf-atoms coincide. 
\begin{proposition}
\label{prop-equiv}
Let $P$ be a basic program. Each answer set by complement of $P$ 
is an answer set by reduct of $P$. 
Furthermore, if $P$ is naf-monotone, then each 
answer set by reduct of $P$ is also an 
 answer set by complement of $P$.
\end{proposition}
The above proposition implies that, in general, the negation-as-failure
by complement approach is more `skeptical' than 
the negation-as-failure by reduct approach, in that it may accept 
fewer answer sets.\footnote{Note that we use the term ``skeptical'' to 
indicate acceptance of fewer models, which is somewhat different than
the use of the term in model theory.}
 Furthermore, Examples \ref{complement} and  \ref{ex5}
show that a minimal (w.r.t. set inclusion) answer set 
by reduct is not necessarily an 
answer set by complement of a program. 

\bigskip

Let $P$ be a normal logic program (without c-atoms)
and let $\catom(P)$ be the 
program obtained by replacing each occurrence of an atom  $a$ in $P$ 
with $(\{a\},\{\{a\}\})$.
Since $(\{a\},\{\{a\}\})$ is a monotone c-atom, 
$\catom(P)$ is a monotone program. 
Proposition~\ref{prop-equiv} implies that, for
$\catom(P)$,  answer sets by reduct and  answer sets
by complement coincide. In the next proposition, 
we prove that the notion of answer sets for programs 
with c-atoms  preserves the notion of answer set for normal 
logic programs, in the following sense. 
\begin{proposition}
[Preserving Answer Sets]
\label{prop-preserv}
For a normal logic program $P$, $M$ is an answer set 
(by complement or by reduct) 
of $\catom(P)$ 
iff $M$ is an answer set of $P$ (w.r.t. the definition in 
\citeA{GelfondL88}).
\end{proposition}
The above proposition, together with Proposition \ref{prop-equiv}, 
implies that normal logic programs can be represented by positive 
basic programs. This is stated in the following corollary. 

\begin{corollary}
  Every answer set of a normal logic program $P$ 
  is an answer set of ${\cal C}(\catom(P))$ and vice versa.
\end{corollary}
 
In the next proposition, we study the minimality and supportedness 
properties of answer sets of basic programs. 
\begin{proposition}
[Minimality of Answer Sets]
\label{prop-min1}
The following properties hold:
\begin{enumerate}
\item Every answer set by complement of a basic program $P$
    is a minimal model of $P$.
\item Every answer set by reduct of a basic, 
    naf-monotone program $P$ is a minimal model of $P$.
\item Every answer set (by complement/reduct) of a basic program 
    $P$ supports 
      each of its members. 
\end{enumerate}
\end{proposition}

\section{Answer Sets for Basic Programs: A Level Mapping Based Approach}\label{secondapp}

The definition of answer sets provided in the previous section 
can be viewed as a generalization of the answer set semantics 
for normal logic programs---in the sense that it  relies on 
a fixpoint operator, defined for positive programs. 
In this section, we discuss another approach for defining 
answer sets for programs with c-atoms, 
which is based on the notion of \emph{well-supported models}.

The notion of well-supported models for normal logic programs was 
introduced in \citeA{Fages94}, and it provides an interesting
alternative characterization of answer sets. Intuitively, a model $M$
of a program $P$ is a well-supported model iff 
there exists a level mapping, from atoms in $M$ 
to the set of positive integers, such that each atom $a \in M$
is supported by a rule $r$, whose body is satisfied by $M$ and the level of 
each positive atom in $body(r)$ is strictly smaller than the level of 
$a$.\footnote{This implicitly means that $pos(r) \subseteq M$ and 
$neg(r) \cap M = \emptyset$, i.e., naf-atoms are dealt with by reduct.} 
Fages proved
that answer sets are well-supported models and vice versa \cite{Fages94}. 
The notion of well-supportedness has been extended to deal with
dynamic logic programs in \citeA{BantiABH05}.
Level mapping has also been used as an effective tool to analyze
different semantics of logic programs in a uniform way \cite{HitzlerW05}. 

In what follows, 
we will show that the notion of 
well-supported models can be effectively applied to programs with c-atoms. 
A key to the formulation of this notion is the answer to the
following question: \\
\centerline{\emph{``what is  the level of a c-atom $A$ given a set of atoms $M$ and 
a level mapping $L$ of $M$?''}} 
On one hand, one might argue that the level mapping of 
$A$ should be defined independently from the mapping of the other atoms, being
$A$ an atom itself. 
On the other hand, it is reasonable to assume that the level of $A$ 
\emph{depends} on 
the levels of the atoms in $A_d$\/, since the satisfaction of $A$ 
(w.r.t. a given interpretation) depends on the satisfaction of the elements 
in $A_d$. 
The fact that every existing semantics of programs with c-atoms 
evaluates the truth value of a c-atom $A$ based on the truth value
assigned to elements of $A_d$ stipulates us to adopt the second view.

It is worth to mention that this view also allows us to avoid
undesirable circular justifications of elements of a well-supported model: if we
follow the first view, the program $P_7$ consisting of the following rules 
\[\begin{array}{lcl}
a & \leftarrow & b\\
b & \leftarrow & a\\
a & \leftarrow & (\{a,b\},\{\emptyset,\{a,b\}\})
  \end{array}
\]
would have $\{a,b\}$ as a well-supported model in 
which $a$, $b$, and $(\{a,b\},\{\emptyset,\{a,b\}\})$ are supported by
$(\{a,b\},\{\emptyset,\{a,b\}\})$, $a$, and $\{a,b\}$
respectively. This means that $a$ is true because {\em both} $a$ and 
$b$ are true, i.e., there is a circular justification for $a$ w.r.t. 
the model $\{a,b\}$. 

\smallskip

Let $M$ be a set of atoms, $\ell$ be a mapping from $M$ to positive integers,  
and let $A$ be a c-atom. We define $H(X) = {\tt max} (\{\ell(a) \mid a \in X\})$, 
and  
$$L(A,M) = {\tt min}(\{ H(X) \mid X \in A_c, X \subseteq M, X \models_M A\}).$$
Intuitively, the ``level'' of each atom is given by the smallest of the levels
of the solutions of the atom compatible with $M$---where the level of a solution
is given by the maximum level of atoms in the solution.
We assume that 
${\tt max}(\emptyset) = 0$, while ${\tt min}(\emptyset)$ is undefined. 
We will now introduce two different notions of well-supported models.
The first notion, called {\em weakly well-supported models}, 
is a straightforward generalization of the definition 
given in \citeA{Fages94}---in that it ignores the naf-atoms. The second
notion,, called {\em strongly well-supported models}, does take into consideration 
the naf-atoms in its definition. 

\begin{definition}
[Weakly Well-Supported Model]
\label{def-well}
Let $P$ be a basic program. A model $M$ of $P$ is 
said to be {\em weakly well-supported} iff 
there exists a level mapping $\ell$ such that,
for each $b \in M$, $P$ contains a rule $r$ with 
$head(r) = (\{b\},\{\{b\}\})$, $M \models body(r)$, and 
for each $A \in pos(r)$, $L(A,M)$ is defined and $l(b) > L(A,M)$.
\end{definition}
We illustrate this definition in the next example. 
\begin{example}
{\rm 
Let us consider again the program $P_5$ and the set of atoms $M = \{a,b\}$. Let 
$A = (\{a,b\},\{\emptyset,\{a,b\}\})$. Obviously, $M$ is a model of $P_5$.
Assume that $M$ is a weakly well-supported model of $P_5$. This means 
that there exists a mapping $\ell$ from $M$ to the set of positive integers
satisfying the condition of Definition \ref{def-well}. 
Since $b \in M$ and there is only one rule in $P_5$ with $b$ as head, we can 
conclude that $\ell(b) > \ell(a)$. 
Observe that $\emptyset \not\models_M A$ and $\{a,b\} \models_M A$. Thus, 
by the definition of $L(A,M)$, we have that 
\[
L(A,M) = {\tt max}(\{\ell(a), \ell(b)\}) =  \ell(b).
\]
This implies that there exists no 
rule in $P_5$, which satisfies the condition of Definition \ref{def-well} 
and has $a$ as its head. In other words, $M$ is not a 
weakly well-supported model of $P_5$. 
}
\hfill$\Box$
\end{example}
The next proposition generalizes Fages' result to answer sets
by reduct for programs with c-atoms.
\begin{proposition} \label{prop-well}
A set of atoms $M$ is an answer set by reduct of
a basic program $P$ iff it is a weakly well-supported model of $P$.
\end{proposition}
As we have seen in the previous section, different ways to 
deal with naf-atoms lead to different semantics for basic programs with c-atoms. 
To take into consideration the fact that naf-atoms can be dealt with 
by complement, we develop an alternative generalization of Fages's definition 
of a well-supported model to programs with abstract c-atoms as follows. 

\begin{definition}
[Strongly Well-Supported Model]
Let $P$ be a basic program. A model $M$ of $P$ is 
said to be {\em strongly well-supported} iff 
there exists a level mapping $\ell$ such that,
for each $b \in M$, $P$ contains a rule $r$ with 
$head(r) = (\{b\},\{\{b\}\})$, $M \models body(r)$, 
for each $A \in pos(r)$, $L(A,M)$ is defined and $\ell(b) > L(A,M)$,
and 
for each $A \in neg(r)$, $L(\bar{A},M)$ is defined and $\ell(b) > L(\bar{A},M)$,
\end{definition}
Using Proposition \ref{prop-well} and Proposition 
\ref{prop-equiv},  we can easily show that 
the following result holds.
\begin{proposition}
A set of atoms $M$ is an answer set by complement of
a basic program $P$ iff it is a strongly well-supported model of ${\cal C}(P)$.
\end{proposition}
The above two propositions, together with Proposition \ref{prop-equiv}, 
lead to the following corollary. 

\begin{corollary} \label{cr-wsp}
For every naf-monotone basic program $P$,  each weakly well-supported 
model of $P$ is also a strongly well-supported model of $P$ 
and vice versa.
\end{corollary}
As we have discussed in the previous section, each 
normal logic program $P$ can be easily translated 
into a monotone basic program with c-atoms of the form $(\{a\},\{\{a\}\})$,
$\catom(P)$. Thus, Corollary \ref{cr-wsp} indicates that the notion of 
weakly/strongly well-supported model is indeed a generalization of 
Fages's definition of well-supported model to programs with c-atoms. 

\section{Answer Sets for General Programs}\label{general}

General programs are programs with non-elementary c-atoms in the head. 
The usefulness of rules with non-elementary c-atoms in the head, in the form 
of a weight constraint or an aggregate,  has been 
discussed in \citeA{Ferraris05a,sim02} and in \citeA{SonPE06}. For example,
a simple atom\footnote{ 
	Recall that aggregates are special form of c-atoms. 
}
\[
\textsc{Count}(\{X \mid taken(X, ai)\}) \le 10
\]
can be used to represent the constraint that no more than 10 students 
can take the AI class. The next example shows how 
the 3-coloring problem of a graph $G$ can
be represented using c-atoms. 


\begin{example}
{\rm
Let the three colors be red ($r$), blue ($b$), and green ($g$). 
The program contains the following rules:

\begin{itemize}
\item the set of atoms $edge(u,v)$ for every edge $(u,v)$ of $G$,

\item for each vertex  $u$ of $G$, the following rule: 
\[
(\{color(u,b), color(u,r), color(u,g)\},
	\{\{color(u,b)\}, \{color(u,r)\}, \{color(u,g)\}\}) \leftarrow
\]
which states that $u$ must
be assigned one and only one of the colors red, blue, or green. 

\item for each edge $(u,v)$ of $G$, three rules representing the
constraint that $u$ and $v$ must have different color:
\begin{eqnarray*}
\bot \la color(u, r), color(v, r), edge(u,v) \\
\bot \la color(u, b), color(v, b), edge(u,v) \\
\bot \la color(u, g), color(v, g), edge(u,v)
\end{eqnarray*}
\end{itemize}}
\qed
\end{example}

We note that, with the exception of the 
proposals in \citeA{Ferraris05a,SonPE06}, other approaches to defining answer 
sets of logic programs with aggregates do not deal with programs with aggregates 
in the head. On the other hand, weight constraint and choice atoms are allowed in the 
head \cite{sim02}. Similarly, c-atoms are considered as head of rules in the framework of 
logic programs with c-atoms by \citeA{MarekR04} and by \citeA{MarekT04}. 

In this section, we define answer sets for general programs---i.e.,
programs where the rule heads can be arbitrary c-atoms. 
Our approach is to convert a program $P$ with c-atoms in the head
into a collection of basic programs, whose answer sets are defined 
as answer sets of $P$. To simplify the presentation,
we will talk about  ``an answer set of a basic program'' to refer 
to either an answer set by complement, an answer set by reduct,
 or a well-supported model of the program. The distinction will 
be stated clearly whenever it is needed.

Let $P$ be a program, $r \in P$, and let $M$ be a model of $P$.
The \emph{instance} of $r$ w.r.t. $M$, denoted by 
$inst(r,M)$ is defined as follows
\[
inst(r,M) = \left\{\begin{array}{lcl}
		\{b \leftarrow body(r)\:|\: b\in M\cap head(r)_d\}
	 & \hspace{.2cm} & M\cap head(r)_d \in head(r)_c\\
		\emptyset & & \textit{otherwise}
		   \end{array}
	\right.
\]

The \emph{instance of $P$ w.r.t. $M$}, denoted by $inst(P,M)$, is the  program
\[inst(P,M) = \bigcup_{r\in P} inst(r,M)\]
It is easy to see that the instance of $P$ w.r.t. $M$ is a basic program.
This allows us to define answer sets of general programs 
as follows. 
\begin{definition} \label{def-as}
Let $P$ be a general program. $M$ is an answer set of $P$ iff 
$M$ is a model of $P$ and $M$ is an answer set of $inst(P,M)$.
\end{definition}
This definition is illustrated in the next examples.
\begin{example}
{\rm 
Let $P_8$ be the program consisting of a single fact:
\[
	(\{a,b\},\{\{a\},\{b\}\}) \leftarrow 
\]
Intuitively, $P_8$ is the choice atom $1 \: \{a,b\} \: 1$ 
in the notation of {\sc Smodels}. 

This program has two models, $\{a\}$ and $\{b\}$. The instance
$inst(P_8,\{a\})$ contains the single fact
\[
	a \leftarrow
\]
whose only answer set is $\{a\}$.
Similarly, the instance $inst(P_8,\{b\})$ is the single fact
\[
	b \leftarrow
\]
whose only answer set is $\{b\}$.
Thus, $P_8$ has the two answer sets $\{a\}$ and $\{b\}$.
}
\hfill $\Box$
\end{example}
The next example shows that in the presence of non-elementary c-atoms
in the head, answer sets might not be minimal. 
\begin{example}
{\rm 
Let $P_9$ be the program consisting of the following rules:
\[
\begin{array}{rcl} 
	(\{a,b\},\{\{a\},\{b\}, \{a,b\}\}) & \leftarrow & \\
	c				   & \leftarrow & b \\
\end{array}
\]
Intuitively, the first rule of 
$P_9$ is the cardinality constraint  $1 \: \{a,b\} \: 2$ 
in the notation of {\sc Smodels}. This program has four models: 
$M_1 =\{a\}$, $M_2=\{b,c\}$, $M_3=\{a,b,c\}$, and $M_4 = \{a,c\}$.
The instance $inst(P_9,M_1)$ contains the single fact
\[ a \leftarrow \]
whose only answer set is $M_1$. 
Thus, $M_1$ is an answer set of $P_9$.

If we consider $M_3$, the corresponding instance $inst(P_9,M_3)$ contains the rules
\[\begin{array}{lcl}
	a & \leftarrow &\\
	b & \leftarrow & \\
	c & \leftarrow & b \\
  \end{array}
\]
whose only answer set is $M_3$. This shows that $M_3$ is another answer 
set of $P_9$. 

Similarly, one can show that $M_2$ is also an
answer set of $P_9$.

The instance $inst(P_9, M_4)$ is the program
\[
\begin{array}{rcl} 
a & \leftarrow & \\
c & \leftarrow & b \\
\end{array}
\]
which has $\{a\}$ as its only answer set. Hence, $M_4$ is not an answer 
set of $P_9$. 
Thus, $P_9$ has three answer sets, $M_1$, $M_2$,
and $M_3$. In particular, observe that $M_1\subset M_3$.
}
\hfill $\Box$
\end{example}

Observe that if $P$ is a basic program then $P$ is its unique
instance. As such, the notion of answer sets for general programs
is a generalization of the notion of answer sets 
for basic programs. It can be shown that Proposition
\ref{prop-equiv} also holds for general 
programs. The relationship between the notion of answer set
for general programs and the definition given in \citeA{MarekR04} and 
other extensions to logic programming is discussed in the next section.

\section{Related Work and Discussion}\label{discuss}

In this section, we relate our work to some recently proposed 
extensions of logic programming, and discuss a possible method
for computing answer sets of programs with c-atoms using 
available answer set solvers.  

\subsection{Related Work}

The concept of \emph{logic programs with c-atoms}, as used
in this paper, has been originally introduced in  \citeA{MarekR04} and in \citeA{MarekT04}---in
particular, programs with c-atoms have been named \emph{SC-programs}
in \citeA{MarekR04}.\footnote{Although naf-atoms are not allowed in the definition of SC-programs,
   the authors  suggest that naf-atoms can be replaced by 
   their complement.} 
The Example~\ref{ex0} shows that our semantical characterization differs from 
the approach adopted in \citeA{MarekR04}. In particular, our approach guarantees
that answer sets for basic programs are minimal, while that is not the case for the approach
described in~\citeA{MarekR04}. Consider another example:

\begin{example}
{\rm
Consider the program $P_{10}$ 
\[\begin{array}{lcl}
a & \leftarrow & \\
c & \leftarrow&\\
d & \leftarrow & (\{a,c,d\}, \: \{ \{a\}, \{a,c,d\}\})
  \end{array}
\]
According to our characterization, this program has only one answer set,
$M_1=\{a,c\}$. If we consider the approach described in~\citeA{MarekR04},
then we can verify that $M_2=\{a,c,d\}$ is an answer set since 
the NSS-reduct of $P_{10}$ with respect to $M_2$ is
\[\begin{array}{lcl}
a &\leftarrow &\\
c & \leftarrow & \\
d & \leftarrow & (\{a,c,d\}, \: \{ \{a\}, \{a,c\},\{a,d\}, \{a,c,d\}\})
  \end{array}
\]
and the least fixpoint of the one-step provability operator is 
$\{a,c,d\}$. 
\hfill$\Box$
}\end{example}
In this type of examples, it seems hard to justify the presence
of $d$ in the answer set of the original program.
We suspect that the replacement of a c-atom by its closure, 
used in the NSS-reduct, might be the reason for the acceptance of 
unintuitive answer sets in \citeA{MarekR04}. The following 
proposition states that our approach is more skeptical 
than the approach of \citeA{MarekR04}.

\begin{proposition} \label{equi-marek-remmel}
Let $P$ be a positive program. If a set of atoms $M$ is an answer set of $P$ 
w.r.t. Definition \ref{def-as} then it is an answer set of $P$ w.r.t. 
\citeA{MarekR04}. 
\end{proposition}

The syntax of \emph{logic programs with c-atoms}, as used
in this paper, is also used in ~\citeA{LiuT05} and in \citeA{LiuT05b}. One of 
the main differences between our work and the work of \citeA{MarekT04}
is that we consider {\em arbitrary c-atoms}, while the
proposal of \citeA{MarekT04} focuses on {\em monotone (and convex) c-atoms}. 
The framework introduced in this paper can be easily extended 
to disjunctive logic programs considered in \citeA{PelovT04}. 

The immediate consequence operator $T_P$ proposed in this
paper is different from the nondeterministic 
one-step provability operator, $T^{nd}_P$, adopted in \citeA{MarekT04},
in that $T_P$ is deterministic and it is applied only to basic positive 
programs. 
In~\citeA{MarekT04} and in \citeA{LiuT05}, the researchers investigate how several
properties of normal logic programs (e.g., strong equivalence) hold
in the semantics of programs with monotone c-atoms of~\citeA{MarekT04}. We
have not directly studied such properties in the context of our 
semantical characterization; nevertheless, 
as we will see later, Proposition \ref{equiv-marek} implies that 
the results proved in \citeA{LiuT05} are immediately applicable
to our semantic characterization for the class of monotone
programs.
We do, however, focus on the use of well-supported 
models and level mapping in studying answer sets for programs 
with c-atoms, an approach that has not been used before for programs
with c-atoms.

We will next present a result that shows that our approach to define 
answer sets for monotone programs coincides with that of \citeA{MarekT04}. 
\begin{proposition} \label{equiv-marek}
Let $P$ be a  monotone program. A set of atoms $M$ is an answer set 
of $P$ w.r.t. Definition \ref{def-as} iff $M$ is a stable model
of $P$ w.r.t. \citeA{MarekT04}.
\end{proposition}
As discussed earlier, c-atoms can be used
to represent several extensions of logic programs, among them weight 
constraints and aggregates. Intuitively, an aggregate atom 
$\alpha$ (see, e.g., \citeR{ElkabaniPS04,FaberLP04})
can be encoded as a c-atom $(D,C)$, where $D$ consists of 
all atoms occurring in the set expression of 
$\alpha$ and $C \subseteq 2^D$ is such that 
every $X \in C$ satisfies $\alpha$ (see Examples \ref{ex2}-\ref{ex3}).
As indicated in \citeA{MarekT04}, many of the previous proposals dealing
with aggregates do not allow aggregates to occur in the head of rules. Here,
instead, we consider programs with c-atoms in the head. 

With regards to naf-atoms, some proposals (see, e.g., \citeR{ElkabaniPS04}) 
do not  allow aggregates to occur in naf-atoms. The proposal in 
\citeA{FaberLP04} treats naf-atoms by complement, although a reduction 
is used in defining the semantics, while
\citeA{Ferraris05a} argues that, under different logics, 
naf-atoms might require different treatments. 

We will now present some propositions which relate our work to 
the recent works on aggregates. We can prove%
\footnote{
  Abusing the notation, we use a single symbol to denote 
  a program in different notations. 
}:
\begin{proposition}
\label{equiv-faber}
For a program with monotone aggregates $P$, $M$ is an answer set 
of $P$ iff it is an answer set of $P$ w.r.t.  
\citeA{FaberLP04} and \citeA{Ferraris05a}. 
\end{proposition}
The proposal presented  in \citeA{Pelov04} and in \citeA{DeneckerPB01} deals 
with aggregates by using
approximation theory and three-valued logic, building
the semantics on the three-valued 
immediate consequence operator $\Phi^{aggr}_P$, which maps three-valued
interpretations into three-valued interpretations of the program. 
This operator can be viewed as an operator which maps pairs of 
set of atoms $(R,S)$ where $R \subseteq S$ into pairs of set of atoms
$(R',S')$ with $R' \subseteq S'$. 
The authors show that the ultimate approximate aggregates provide the 
most precise semantics for logic programs with aggregates. 
Let us denote 
with $\Phi^{1}(R,S)$ and $\Phi^{2}(R,S))$ the two components of 
$\Phi^{aggr}_P(R,S)$, i.e., 
$\Phi^{aggr}_P(R,M) = (\Phi^{1}(R,M), \Phi^{2}(R,M))$. 
The next proposition relates $T_P$ to $\Phi^{aggr}_P$.
\begin{proposition}\label{equiv-pelov}
Let $P$ be a positive program with aggregates and $R$ 
and $M$ be two set of atoms such that $R \subseteq M$. Then, 
$T_P(R,M) = \Phi^{1}(R,M)$. 
\end{proposition}
The above proposition, together with the fact that 
the evaluation of the truth value of aggregate formulas 
in \citeA{DeneckerPB01} treats naf-atoms by complement, allows us 
to conclude that, for a program with aggregates $P$, 
answer sets by complement of $P$ (w.r.t. Definition \ref{basic-pos-a.s.}) 
are ultimate stable models of $P$ \cite{DeneckerPB01} and vice versa.
This result, together with the results in \citeA{SonP05}, 
allows us to conclude that $T_P$ is a generalization of 
the immediate consequence operator for aggregates programs in 
\citeA{SonP05}.

Before we conclude the discussion on related work, we would like to 
point out that Propositions \ref{equi-marek-remmel}-\ref{equiv-pelov} 
show that the different approaches to dealing with aggregates
differ only for non-monotone programs. The main difference 
between our approach and others lies in the skepticism of 
the $T_P$ operator, caused by the notion of conditional 
satisfaction. We will illustrate this issue in the next two examples. 

\begin{example} \label{ex19}
{\rm
Consider the program $P_2$ of Example \ref{ex3}.
This program does not have an answer set 
w.r.t. Definition \ref{basic-pos-a.s.} but has $M = \{p(1), p(-1), p(2)\}$
as an answer set according to \citeA{MarekR04}. The reason for 
the unacceptability of $M$ as an answer set in our approach lies
in that the truth value of the 
aggregate atom $\textsc{Sum}(\{X \mid p(X)\})$ could be either true 
or false even when $p(1)$ is known to be true. This prevents the 
third rule to be applicable and hence the second rule as well.
This makes $p(1)$ the fixpoint of the $T_{P_2}$ operator, given
that $M$ is considered as an answer set. In other words, we cannot
regenerate $M$ given the program---and the skepticism of 
$T_{P_2}$ is the main reason. We observe that other approaches 
(see, e.g., \citeR{FaberLP04,Ferraris05a}) do not accept $M$ as an 
answer set of $P_2$ as well. }
\qed 
\end{example}
The following example shows the difference between our approach 
and those in \citeA{FaberLP04} as well as in \citeA{Ferraris05a}.

\begin{example}
\label{exp9}
{\rm
Consider the program $P$ 
\[
\begin{array}{lll}
p(1) & \leftarrow & (\{p(1),p(-1)\},\{\emptyset, \{p(1),p(-1)\}\}) \\ 
p(1) & \leftarrow & p(-1) \\
p(-1)& \leftarrow & p(1) \\
\end{array}
\]
Intuitively, the abstract atom 
$A = (\{p(1),p(-1)\},\{\emptyset, \{p(1),p(-1)\}\})$
represents the aggregate atom 
$\textsc{Sum}(\{X \mid p(X)\}) \ge 0$. 
This program has two models $M_1 = \{p(1), p(-1)\}$ 
and $M_2 = \emptyset$. The approaches in 
\citeA{MarekR04,FaberLP04}, and \citeA{Ferraris05a} accept $M_1$ as an answer set, 
while our approach and that of \citeA{Pelov04,DeneckerPB01} do not admit 
any answer sets. In our approach, 
$T_P(\emptyset,M_1) = \emptyset$ because $\emptyset$ does not conditionally
satisfy $A$ w.r.t $M_1$ since it is not true in every possible 
extension of $\emptyset$ that leads to $M_1$, namely it is not true 
in $\{p(-1)\}$. In other words, the skepticism of our approach 
is again the main reason for the difference between our approach 
and the approaches in \citeA{FaberLP04} and in \citeA{Ferraris05a}.}
\hfill$\Box$
\end{example}

\subsection{Discussion}
In this section, we briefly discuss a possible method for computing 
answer sets of programs with c-atoms, using off-the-shelf answer set 
solvers. The method makes use of a transformation similar to the 
unfolding transformation proposed in \citeA{ElkabaniPS04} for dealing with aggregates, 
which has been further studied and implemented in \citeA{ElkabaniPS05}. 

We begin our discussion with basic 
positive programs. 
Given a basic positive program $P$ and a c-atom $A$, if $A_c \ne \emptyset$, 
an unfolding of $A$
is an expression of the form 
\[
p_1,\ldots,p_n,\naf q_1,\ldots,\naf q_m
\]
where 
$\{p_1,\ldots,p_n\} \in  A_c$ and 
$\{q_1,\ldots, q_m\} = A_d \setminus \{p_1,\ldots,p_n\}$. If $A_c = \emptyset$, then 
$\bot$, denoting {\em false},  is the unique unfolding of $A$. 
Observe that if $A = (\{a\},\{\{a\}\})$ then the only unfolding of $A$ is $a$. 
An unfolding of a rule 
\[
A_0 \leftarrow A_1,\ldots,A_k
\]
is a rule obtained by replacing each $A_i$ with one of 
its unfoldings. $unfolding(r)$ denotes the set of all the unfoldings of a rule $r$. Let 
$unfolding(P) = \bigcup_{r \in P} unfolding(r)$. Clearly, 
$unfolding(P)$ is a normal logic program if $P$ is a basic positive program. 
We can show that $M$ is an 
answer set of $P$ iff $M$ is an answer set of $unfolding(P)$. 
This indicates that we can compute answer sets of basic positive 
programs with c-atoms by ({\em i}) computing 
its unfolding; and ({\em ii}) using available answer set solvers to 
compute the answer sets of the unfolded program. Following this approach,
the main additional cost for computing answer sets of a basic positive program
is the cost incurred during the unfolding process. Theoretically, this can be 
very costly as for each rule $r$, we have that $|unfolding(r)| = \Pi_{A \in body(r)} |A_c|$,
where $|.|$ denotes the cardinality of a  set. This means that 
the size of the program $unfolding(P)$ might be exponential in the size
of the original program $P$. Thus, the additional cost might be significant. 
In practice, we can expect that this number is 
more manageable, as a rule might contain only a few c-atoms whose set of 
solutions is reasonably small. Furthermore, certain techniques can be employed to
reduce the size of the unfolding program \cite{SonPE06}. 

The above method can be easily extended to deal with naf-atoms and general programs. 
If answer sets by complement need to be computed, we need to 
({\em i}) compute the complement of the program;
and ({\em ii}) use the above procedure to compute answer sets of
the complement. 
On the other hand, if answer sets by reduct need to be computed, we will have at 
hand a tentative answer set $M$. The reduction of the program with respect to 
$M$ can be computed, and the unfolding can then be applied to verify whether $M$ 
is an answer set of the reduct. Observe that the complement or a reduct of 
a program can be easily computed, and it does not increase the size of the program.
As such, the main cost for computing answer sets of general programs
following this approach is still the cost of the unfolding.
So far, in our study on programs with aggregates (a special type of c-atoms), we
did not encounter unmanageable situations \cite{SonPE06}.

Observe that the specification of a c-atom requires the enumeration of 
its domain and solutions, whose size can be exponential in the size of the
set of atoms of the program. This does not mean that an explicit 
representation of c-atoms needs to be used. In most cases, c-atoms can 
be replaced by aggregate literals. Because of this, several complexity 
results for 
programs with aggregates (see, e.g., \citeR{Pelov04,SonP05}) can be extended
to logic programs with c-atoms. For example, we can easily show that 
the problem of determining whether a logic program has an answer 
set or not is at least NP$^{\textnormal{co{-}NP}}$. However, for programs 
with c-atoms representable by standard aggregate functions, 
except those of the form $\textsc{Sum}(.) \ne value$ 
and $\textsc{Avg}(.) \ne value$, the problem of determining 
whether or not a program has an answer set
remains NP-complete. 

\section{Conclusions and Future Work}\label{conc}

In this paper, we explored
a general logic programming
framework based on the use of arbitrary constraint atoms. The proposed
approach provides a characterization which is more in line with existing 
semantics of logic programming with aggregates than
the characterization proposed in~\citeA{MarekR04}.
We provided two alternative characterizations of 
answer set semantics for programs with arbitrary constraint atoms,
the first based on a fixpoint operator, which generalizes the immediate consequence
operator for traditional logic programs, and the
second built on a generalization of the notion of well-supported 
models of \citeA{Fages94}. 

Within each characterization of answer set, we investigated two 
methodologies for treating naf-atoms and 
we identified the class of naf-monotone programs, on which the two 
approaches for dealing with naf-atoms coincide. 
We also proved that the newly proposed semantics coincides 
with the semantics proposed in \citeA{MarekT04} for 
monotone programs. Finally, we related our work to other
proposals on logic programs with aggregates and discussed a
possible method for computing answer sets of programs with abstract 
constraint atoms using available answer set solvers. 

The proposal has some unexplored aspects.
The proposed
approach is rather ``skeptical'' in the identification of answer 
sets---while the approach in~\citeA{MarekR04} is overly 
``credulous''. We believe that these two approaches represent the two
extremes of a continuum that needs to be explored. In particular,
we believe it is possible to identify ``intermediate'' approaches
simply by modifying the notion of conditional satisfaction. Work is
in progress to explore these alternatives.

\subsection*{Acknowledgment} 

The authors wish to thank the anonymous
reviewers for their insightful comments.
The research has been partially supported by NSF grants
HRD-0420407, CNS-0454066, and CNS-0220590.
An extended abstract of this paper appeared in the Proceedings of the  
Twenty-First National Conference on Artificial Intelligence, 2006. 


\section*{Appendix A}
First, we will show some lemmas that will be used for the proofs
of propositions.

\begin{lemma}
\label{lm-cond}
Let $S \subseteq U \subseteq M' \subseteq M$ be sets of atoms and $A$ be an 
abstract constraint atom. Then, $S \models_M A$ implies $U \models_{M'} A$.
\end{lemma}
{\bf Proof.}
$S \models_M A$ implies that $$\{I \mid I \subseteq A_d,
S \cap A_d \subseteq I \subseteq M \cap A_d\} \subseteq A_c.$$ 
Together with the fact $S \subseteq U \subseteq M' \subseteq M$, we have that 
$$\{I \mid I \subseteq A_d,
U \cap A_d \subseteq I \subseteq M' \cap A_d\} \subseteq A_c.$$
This implies that $U \models_{M'} A$.
\qed

\begin{lemma}
\label{lm5}
For two sets of atoms $M' \subseteq M$ and a monotone c-atom $A$, 
if $M' \models A$ (resp. $M' \models \bar{A}$) and 
$M \models A$ (resp. $M \models \bar{A}$) then
$M' \models_M A$ (resp. $M' \models_M \bar{A}$).
(Recall that $\bar{A}$ denotes the complement of $A$.)
\end{lemma}
{\bf Proof. }
\begin{enumerate}
\item Let us assume that  $M' \models A$ and $M \models A$. From the
    monotonicity of $A$, we can conclude that, for every $S$, if $M' \subseteq
    S \subseteq M$, we have that $S \models A$. As a result, we
have $M' \models_M A$.
\item Let us assume that $M' \models \bar{A}$ and $M \models \bar{A}$.
    Let us assume, by contradiction, that $M' \not \models_M \bar{A}$. 
    Since we already know that $M'\models \bar{A}$, this implies that there exists
    $S\subseteq A_d$, $M'\cap A_d \subseteq S \subseteq M\cap A_d$, such that 
    $S \not\in 2^{A_d}\setminus A_c$, i.e., $S \in A_c$.
    Since $A$ is monotone and $S \subseteq M$,
    we have that  $M \models A$. This is a contradiction, since we 
    initially assumed that $M \models \bar{A}$.
\end{enumerate}
\qed


\medskip \noindent
{\bf Proposition \ref{prop-tp}.}
Let $M$ be a model of $P$, and let 
$S \subseteq U \subseteq M$. Then
$$T_P(S, M) \subseteq T_P(U,M) \subseteq M.$$

\smallskip
\noindent
{\bf Proof.}
\begin{enumerate}
\item From Lemma \ref{lm-cond}, the assumption that $S \subseteq U
\subseteq M$, and the definition of $T_P$, we have that 
 $T_P(S,M) \subseteq T_P(U,M)$.
 
\item Let us now show that $T_P(U,M) \subseteq M$. Consider an atom $a 
\in T_P(U,M)$. We need to show that $a \in M$. From the definition of 
the $T_P$ operator, there
is a rule $r$ such that $head(r)=(\{a\},\{\{a\}\})$ and $U \models_M pos(r)$. But observe
that, for each $A \in pos(r)$, if $U \models_M A$ then we will
have that $M \models A$ (Definition \ref{cond-sat}).
Thus, we can conclude that $M \models pos(r)$. Since the program
is positive and $M$ is known to be a model of $P$, we must have
that $M\models head(r)$, thus $a \in M$.
\end{enumerate}
\hfill $\Box$


\medskip\noindent
{\bf Corollary~\ref{cr-minm}}
Let $P$ be a positive basic program and $M$ be an answer set of $P$. 
Then, $M$ is a minimal  model of $P$.

\smallskip
\noindent
{\bf Proof. }
$M$ is a model of $P$ since it is an answer set of $P$ (Definition \ref{basic-pos-a.s.}). 
Thus, we need to prove that $M$ is indeed a \emph{minimal}
model of $P$. Suppose that there exists
$M' \subset M$ such that $M'$ is a model of $P$.
Proposition \ref{prop-tp} and Lemma \ref{lm-cond} 
imply that $T^k(\emptyset,M) \subseteq T^k(\emptyset,M') \subseteq M'$
for every $k$. Since $M$ is an answer set, we have
that $M \subseteq M'$. This contradicts the assumption that 
$M' \subset M$. 
\qed

\bigskip
\noindent
{\bf Proposition~\ref{prop-equiv}.}
Let $P$ be a basic  program. Each answer set by complement of $P$ 
is an answer set by reduct of $P$. 
Furthermore, if $P$ is naf-monotone, then each 
an answer set by reduct of $P$ is also an 
 answer set by complement of $P$.

\smallskip
\noindent
{\bf Proof.}
Let us start by showing that answer sets by complement are
also answer sets by reduct. Let $M$ be a model and let
us denote with $P_1 = {\cal C}(P)$ and let
$P_2 = P^M$. Using the fact that if $S \models_M \bar{A}$ 
then $M \not\models A$ we can easily prove by induction that 
the following result holds:
\begin{equation}\label{preq1}
T_{P_1}^\infty(\emptyset,M) \subseteq T_{P_2}^\infty(\emptyset,M)
\end{equation}
and if $P$ is a naf-monotone program then 
\begin{equation}
T^i_{P_1}(\emptyset,M) = T^i_{P_2}(\emptyset,M) \label{eq-1}
\end{equation}
If $M$ is an answer set by complement then we have
$M = T_{P_1}^\infty(\emptyset,M)$. Furthermore, 
$T_{P_2}^\infty(\emptyset,M) \subseteq M$ (Proposition \ref{prop-tp}). 
This implies that $M$ is an answer set of $P$ by reduct as well.

If $P$ is naf-monotone, using Equation (\ref{eq-1}) and the fact 
that $M$ is an answer set of $P_2$ we can conclude that $M$ is 
an answer set by complement of $P$. 
\qed

\bigskip

\noindent
{\bf Proposition~\ref{prop-preserv}.}
For a normal logic program $P$, $M$ is an answer set 
(by complement or by reduct) 
of $\catom(P)$ 
iff $M$ is an answer set of $P$ (w.r.t. Definition in 
\citeA{GelfondL88}).

\smallskip
\noindent
{\bf Proof.}
For convenience, in this proof, we will refer 
to answer sets defined in \citeA{GelfondL88}
as \emph{GL-answer sets}. 
Because of the monotonicity of $\catom(P)$ and Proposition 
\ref{prop-equiv}, it suffices to show that $M$ is 
an answer set of $P$ iff $M$ is an answer set by reduct of $\catom(P)$.

Let us consider the  case where  $P$ is a positive program. 
It follows from Observation \ref{observ1} and the fact that
$S \models_M (\{a\},\{\{a\}\})$ iff $a \in S$ that the operator 
$T_P(.,.)$ for $P$ (defined in Observation \ref{observ1}) coincides
with the operator $T_P(.,.)$ for $\catom(P)$. Hence, 
$M$ is an answer set of $\catom(P)$ iff 
$M$ is a GL-answer set of $P$. 

Now suppose that $P$ is an arbitrary normal logic program.
Let $GL(P,M)$ be the Gelfond-Lifschitz's reduct of $P$ w.r.t. 
$M$. Since $M \models (\{a\},\{\{a\}\})$ iff $a \in M$,
we have that $\catom(P)^M = \catom(GL(P,M))$. Using the result for 
positive program, we have that 
$M$ is a GL-answer set of $P$ iff $M$ is an answer set 
by reduct of $\catom(P)$.
\qed

\bigskip
\noindent
{\bf Proposition~\ref{prop-min1}.}
\begin{enumerate}
\item Every answer set by complement of a basic program $P$
    is a minimal model of $P$.
\item Every an answer set by reduct of a basic, 
    naf-monotone program $P$ is a minimal model of $P$.
\item Every answer set (by complement/reduct) of a basic program 
    $P$ supports 
      each of its members. 
\end{enumerate}
\noindent
{\bf Proof.}

\begin{enumerate}

\item 
Notice that for a set of atoms $M$  and an abstract constraint
atom $A$, $M \models \naf A$ iff $M \models \bar{A}$. This implies
that $M$ is a model of $P$ iff $M$ is a model of ${\cal C}(P)$. From 
this and from Corollary \ref{cr-minm}, we can conclude that
if $M$ is an answer set  by complement of $P$, then it is a 
minimal model of $P$.

\item Let us assume that $P$ is a naf-monotone basic program,
and let  $M$ be an answer set by reduct of $P$. From 
Proposition \ref{prop-equiv}, $M$ is
also an answer set of $P$ by complement.
The previous result implies that $M$ is a minimal model of $P$.

\item It follows from Proposition \ref{prop-equiv} that 
    it is enough to prove the conclusion for answer sets by reduct of $P$. 
    Let $M$ be an answer set by reduct of $P$. From
    the definition, we have that $M = T^{\infty}_{P^M}(\emptyset,M)$. This
    implies that, if $a\in M$, then there exists $i$ such that
    $a \in T^i_{P^M}(\emptyset,M)$. In turn, this allows us to identify
    a rule 
\[ (\{a\},\{\{a\}\}) \leftarrow A_1, \dots, A_k, \naf A_{k+1},\dots, \naf A_{n} \]
    such that $M\not\models A_j$ for $k+1\leq j \leq n$ and 
    $T^i_{P^M}(\emptyset,M) \models_M A_i$ for $1\leq i \leq k$. 
    In particular, $M\models A_i$ for $1\leq i \leq k$. 
    We can easily conclude that the given rule supports $a$.
\end{enumerate}
\qed

\subsection*{Proof of Proposition \ref{prop-well}}
Let us start by proving the following lemma:
\begin{lemma}
\label{lm-well1}
Let $P$ be a positive, basic program, and let  $M$ be a
weakly well-supported model of $P$. Let $l$ be a mapping that satisfies the 
desired properties of weakly well-supportedness of $M$. For every atom $a$, 
$a \in M$ implies $a \in T_P^{l(a)+1}(\emptyset,M)$.
\end{lemma}
{\bf Proof. }
First, observe that, for an atom $a \in M$, 
we have $$L((\{a\},\{\{a\}\}),M) = l(a).$$
Let us prove the lemma by induction on $l(a)$.

\begin{enumerate}
\item \emph{Base Case:} Consider $a \in M$ such that 
$l(a) = 0$. 
Clearly, we have that $P$ must contain the rule
\[ (\{a\},\{\{a\}\}) \leftarrow \]
hence, $a \in T_P^1(\emptyset,M)$.
\item \emph{Inductive Step:} Assume that 
the result holds for every atom $b$
such that $0 \le l(b) < k$\/.

Consider an atom $a \in M$ such that $l(a) = k$. We will show
that $a \in T_P^{k+1}(\emptyset,M)$. Since  $M$ is a weakly 
well-supported model of $P$, 
there exists a rule 
$$(\{a\},\{\{a\}\}) \leftarrow A_1,\dots, A_n$$
in $P$ such that $L(A_i,M)$ is defined and $l(a) > L(A_i,M)$ for 
every $1 \le i \le n$.

Let $S = T_P^{k}(\emptyset,M)$. For each $i \in \{1,\dots,n\}$,
since $L(A_i,M)$ is defined, there is $X \subseteq M$, 
$X \in (A_i)_c$ such that $X \models_M A_i$ and $L(A_i,M) = H(X)$. 
Hence, we have
$k = l(a) > L(A_i,M) = H(X)$. From the inductive hypothesis,
since $X \subseteq M$, we can conclude that  $X \subseteq S$.
On the other hand, we already proved (Corollary \ref{cr-tp}) that 
$$T_P^0(\emptyset,M) \subseteq \dots \subseteq T_P^k(\emptyset,M)=S 
\subseteq \dots \subseteq  M.$$  
As a result, we have that
$X \subseteq S \subseteq M$.

   From Lemma \ref{lm-cond}, since $X \models_M A_i$,
this implies that $S \models_M A_i$. 
Accordingly, we have $a \in T_P^{k+1}(\emptyset,M)$.
\end{enumerate}
\qed

\medskip
We can now proceed with the proof of the proposition.

\medskip \smallskip
\noindent
{\bf Proposition~\ref{prop-well}.}
A set $M$ of atoms is an answer set by reduct of
a basic program $P$ iff it is a weakly well-supported model of $P$.

\medskip
\noindent
{\bf Proof.}
We will prove the proposition in two steps. 
We first prove that the result holds for positive programs and 
then extend it to the case of arbitrary basic programs.

\begin{itemize}
\item 
{\bf $P$ is a positive program.}
\begin{enumerate}
\item ``$\Rightarrow$'': Suppose $M$ is an answer set of $P$. 
Corollary~\ref{cr-minm} implies that $M$ is a model of $P$. Thus, it suffices to 
find a level mapping satisfies $l$ the condition of Definition \ref{def-well}.
For each atom $a$, let 
$$l(a) = \left \{
\begin{array}{ll}
{\tt min} \{k \mid a \in T_P^{k}(\emptyset,M) \} & \textnormal{ if }
a \in M \\
0 & \textnormal{otherwise}
\end{array}
\right.
$$
Clearly, $l$ is well defined. We will show that $l$ is indeed the mapping 
satisfying the properties of Definition \ref{def-well}.

Let us consider an atom $a \in M$ and let $k = l(a)$. Clearly, $k > 0$
since $T_P^{0}(\emptyset,M) = \emptyset$. So, we have that 
$a \in T_P^{k}(\emptyset,M)$ but $a \not \in S=T_P^{k-1}(\emptyset,M)
\subseteq M$. There are two cases:

\begin{enumerate}
\item $P$ contains a rule $$(\{a\},\{\{a\}\}) \leftarrow$$
In this case, the condition on $l$ for the atom $a$ is trivially satisfied. 

\item $P$ contains a rule $r$ of the form 
$$(\{a\},\{\{a\}\}) \leftarrow A_1,\dots,A_n$$ such that $S \models_M A_i$
for $1 \le i \le n$. 

Consider an integer $1 \le i \le n$. Let $X = S \cap (A_i)_d$.
By the definition of conditional satisfaction, we have 
that $X \in (A_i)_c$. It is easy to check that $X \models_M A_i$. 
In addition, we have $X \subseteq M$.
As a result, $L(A_i,M)$ is defined. Furthermore,
we have $L(A_i,M) \le H(X)  \le H(S) < k = l(a)$.
This shows that the condition on $l$ for $a$ is also satisfied in this 
case. 
\end{enumerate}
The above two cases allow us to conclude that $l$ satisfies the condition
of Definition \ref{def-well}, i.e., $M$ is a weakly well-supported model of $P$.

\item ``$\Leftarrow$'': Suppose $M$ is a weakly well-supported model of $P$.
Due to Lemma \ref{lm-well1}, we have that $M \subseteq
T_P^{\infty}(\emptyset,M)$. On the other hand, from Corollary \ref{cr-tp},
we have $T_P^{\infty}(\emptyset,M) \subseteq M$. Consequently,
we have $M = T_P^{\infty}(\emptyset,M)$, which implies
that $M$ is an answer set of $P$.
\end{enumerate}

\item 
{\bf $P$ is an arbitrary basic program.}
It is easy to see that a set of atoms $M$ is a weakly well-supported model
of $P$ iff $M$ is a weakly well-supported model of $P^M$. From the previous
result, this means that $M$ is an answer set by reduct of $P$ iff $M$
is a weakly well-supported model of $P$.
\end{itemize}
\qed

\bigskip
\noindent 
{\bf Proposition~\ref{equi-marek-remmel}.}
Let $P$ be a positive program. If a set of atoms $M$ is an answer set of $P$ 
w.r.t. Definition \ref{def-as} then it is an answer set of $P$ w.r.t. 
\citeA{MarekR04}.

\smallskip
\noindent
{\bf Proof.}
\begin{itemize}
\item Consider the case that $P$ is a basic program. 
Since $NSS(P,M)$ is a monotone positive programs, the least fixpoint 
of the one-step provability operator $T_{NSS(P,M)}(.)$ coincides with 
the least fixpoint of our
extended immediate consequence operator $T_{NSS(P,M)}(.,.)$ 
(see also Proposition \ref{equiv-marek}). Furthermore, 
we can easily verify that 
$T_P^\infty(\emptyset, M) = T_{NSS(P,M)}^\infty(\emptyset, M)$ 
holds if $M$ is an 
answer set w.r.t. Definition \ref{def-as}. These two observations
imply the conclusion of the proposition. 

\item
We now consider the case that $P$ is general positive program.
Without loss of generality, we can assume that $P$ does not 
contain any constraints. 
Let $Q = inst(P,M)$. We have that 
for a rule $r' \in Q$ if and only if there exists some rule $r \in P$ 
such that $M \models head(r)$, 
$r' = a \leftarrow body(r)$, and $a \in head(r)_d \cap M$. 
This implies that $NSS(P,M) = NSS(Q,M)$. Since 
$M$ is an answer set of $Q$ (w.r.t. Definition \ref{def-as}),
we conclude that it is also an answer set of $Q$ w.r.t.
\citeA{MarekR04} (the basic case) which implies that 
$M$ is also an answer set of $P$ w.r.t. \citeA{MarekR04}
\end{itemize}
\qed

\bigskip 
\noindent
{\bf Proposition~\ref{equiv-marek}.}
Let $P$ be a monotone program. A set of atoms $M$ is an answer set 
of $P$ w.r.t. Definition \ref{def-as} iff $M$ is a stable model
of $P$ w.r.t. \citeA{MarekT04}.

\smallskip
\noindent
{\bf Proof.}
Let us start by showing the validity of the result for
positive programs. Let us assume that $P$ is a positive program. 
Without loss of generality, we assume that $P$ does not contain 
any constraints. 

\begin{enumerate}
\item ``$\Rightarrow$'': Let $M$ be an answer set of $P$. From 
Definition~\ref{def-as}, we have that $M$ is a model of $P$
and $M$ is  an answer set of $Q = inst(P,M)$.

For every non-negative integer $i$, let
$$M_i = T_Q^i(\emptyset,M)$$ 
Because $M$ is an answer set of $Q$, by definition, we have 
$$M = T_Q^\infty(\emptyset,M)$$
To show that
$M$ is a stable model of $P$ w.r.t. \citeA{MarekT04}, all we need to do
is to prove that the sequence $\langle M_i \rangle_{i=0}^\infty$ is 
a $P{-}computation$. We do so by proving that (i)
$M_i \subseteq M_{i+1}$ and (ii) for every $r \in P(M_i)$
\footnote{Recall that $P(M_i)$ is the set of rules in $P$ 
whose body is satisfied by $M_i$.}, $M_{i+1} \models head(r)$, and (iii)
$M_{i+1} \subseteq hset(P(M_i))$. 

\begin{enumerate}
\item[(i)] Follows from Corollary \ref{cr-tp}.
\item[(ii)] 
Consider a rule $r \in P(M_i)$. By the definition of
$P(M_i)$, we have that $M_i \models body(r)$. Because $P$ is monotone 
and $M_i \subseteq M$, it follows that $M \models body(r)$
and $M_{i} \models_M body(r)$.

Let $X = M \cap head(r)_d$. By the definition of $inst(r,M)$, we have
that $$inst(r,M) = \{ b \leftarrow body(r) \mid b \in X\} \subseteq Q$$
As $M_i \models_M body(r)$, for every $r' \in inst(r,M)$,
$M_i \models_M body(r')$. By the definition of $M_{i+1}$,
it follows that $head(r') \in M_{i+1}$. Hence, $X \subseteq M_{i+1}$. 
Since $X \subseteq head(r)_d$, this implies that
$$X \subseteq M_{i+1} \cap head(r)_d$$
On the other hand, because $M_{i+1} \subseteq M$, we have
$$M_{i+1} \cap head(r)_d \subseteq X$$
Accordingly, we have
\begin{equation}
M \cap head(r)_d = X = M_{i+1} \cap head(r)_d \label{eq-equiv-marek:eq1}
\end{equation}

On the other hand, because $M$ is a model of $P$ and 
$M \models body(r)$, we have $M \models head(r)$. Therefore,
\begin{equation}
M \cap head(r)_d \in head(r)_c \label{eq-equiv-marek:eq2}
\end{equation}
    From (\ref{eq-equiv-marek:eq1}) and 
(\ref{eq-equiv-marek:eq2}), we have $M_{i+1} \cap head(r)_d
\in head(r)_c$, i.e., $M_{i+1} \models head(r)$. 
\item[(iii)]
Let $a$
be an atom in $M_{i+1}$. From the definition of $M_{i+1}$
it is easy to see that $Q$ must contain a rule $r'$
whose head is $a$ and whose body is satisfied by $M_i$.
This implies that $P(M_i)$ must contain a rule $r$ 
such that $a \in M \cap head(r)_d$. It follows that 
$a \in head(r)_d \subseteq hset(P(M_i))$. Accordingly, we have $M_{i+1}
\subseteq hset(P(M_i))$.
\end{enumerate}

\item ``$\Leftarrow$'': Let $M$ be a stable model  of $P$
according to~\citeA{MarekT04} and let $\langle X_i \rangle_{i=0}^{\infty}$ be 
the canonical computation for $M$, i.e.,
$$X_0 = \emptyset$$
$$X_{i+1} = \bigcup_{r \in P(X_i)} head(r)_d \cap M$$
According to Theorem 5 of \citeA{MarekT04}, we have
$$M = \bigcup_i {X_i}$$

Let $Q = inst(P,M)$. Because $M$ is a stable model of $P$,
it is also a model of $P$. So, to prove that $M$ is an answer
set of $P$, we only need to show that it is an answer set of
$Q$. 

Let us construct a sequence of sets of atoms 
$\langle M_i \rangle_i^{\infty}$ as follows
$$M_0 = \emptyset$$
$$M_{i+1} = T_Q(\emptyset,M_i)$$
Clearly, to prove $M$ is an answer set of $Q$, it suffices
to show that
\begin{equation}
X_i = M_i \label{eq-equiv-marek:eq3}
\end{equation}
Let us prove this by induction.
\begin{enumerate}
\item $i = 0$: trivial because $X_0 = M_0 = \emptyset$.
\item Suppose (\ref{eq-equiv-marek:eq3}) is true for $i = k$, 
we will show that it is true for $i = k+1$.

Consider an atom $a \in X_{k+1}$. By the definition of
$X_{k+1}$, there exists a rule $r \in P$ such that
$a \in head(r)_d$ and $X_k \models body(r)$.
Since $X_k \subseteq M$ and $P$ is monotone,
it follows that $M \models body(r)$. Because a stable
model of $P$ is also a model of $P$, we have
$M \models head(r)$. As a result, $Q$ contains
the following set of rules:
$$inst(r,M) = \{ b \leftarrow body(r) \mid b
\in M \cap head(r)_d\}$$
Because $a \in head(r)_d$ and $a \in X_{k+1} \subseteq M$, 
we have $a \in M \cap head(r)_d$. As a result,
the following rule belongs to $inst(r,M)$
$$a \leftarrow body(r)$$
Because $M_k = X_k$ (inductive hypothesis), we have $M_k \models body(r)$
and thus $M_k \models_M body(r)$ (recall that $M_k = X_k
\subseteq M$ and $body(r)$ consists of monotone abstract
constraint atoms only). By the definition of $M_{k+1}$,
we have $a \in M_{k+1}$.

We have shown that for every atom $a \in X_{k+1}$, 
$a$ belongs to $M_{k+1}$. Hence, 
\begin{equation}
X_{k+1} \subseteq M_{k+1} \label{eq-equiv-marek:eq4}
\end{equation}
Now, we will show that $M_{k+1} \subseteq X_{k+1}$. Consider
an atom $b$ in $M_{k+1}$. By definition of $M_{k+1}$,
there exists a rule $r' \in Q$ such that
$head(r)_d = b$ and $M_k \models_M body(r')$. By the definition
of $Q$ this means that there exists a rule $r$ in $P$ such
that $M \models head(r)_d$, $body(r) = body(r')$ and
$b \in M \cap head(r)_d$. Because $X_k = M_k$ (inductive hypothesis), 
from $M_k \models_M body(r') = body(r)$, we have
$X_k \models body(r)$. This implies that $r \in P(X_k)$.
Hence, 
$$b \in M \cap head(r)_d \subseteq X_{k+1}$$
Therefore we have
\begin{equation}
M_{k+1} \subseteq X_{k+1} \label{eq-equiv-marek:eq5}
\end{equation}
    From (\ref{eq-equiv-marek:eq4}) and (\ref{eq-equiv-marek:eq5}),
we have $X_{k+1} = M_{k+1}$.
\end{enumerate}
\end{enumerate}
The above result can be easily extended for programs with 
negation-as-failure c-atoms. We omit the proof here. 
\qed

\bigskip
\noindent
{\bf Proof of Proposition \ref{equiv-faber}.}

\medskip

To prove this proposition, a brief review of the approach in \citeA{FaberLP04} 
is needed. The notion of answer set proposed  in \citeA{FaberLP04}
is based on a new notion of reduct, defined as follows. 
Given a program $P$ and a set of 
atoms $S$, the {\em reduct of P with respect to S}, denoted by 
${^S}P$, is obtained by removing from $P$ those rules whose 
body is not satisfied by $S$. In other words,
\[
FLP(P,M) = \{r \mid r \in ground(P), S \models body(r)\}.
\]
The novelty of this reduct is that it 
\emph{does not} remove aggregate atoms
and negation-as-failure literals satisfied by $S$. 
For a program $P$, $S$ is a {\em FLP-answer set} of $P$ 
iff it is a minimal model of $FLP(P,S)$. We will now continue with 
the proof of the proposition. It is easy to see that it is enough 
to consider programs without negation-as-failure c-atoms. 

\smallskip
\noindent
{\bf Proposition \ref{equiv-faber}.}
For a program with monotone aggregates $P$, $M$ is an answer set 
of $P$ iff it is an answer set of $P$ w.r.t.  
\citeA{FaberLP04} and \citeA{Ferraris05a}. 

\medskip
\noindent 
{\bf Proof.} Due to the equivalent result in \citeA{Ferraris05a}, it 
suffices to prove the equivalence between our approach and that of 
\citeA{FaberLP04}. Notice that in this paper we are dealing with 
ground programs and therefore  
\begin{enumerate}
\item ``$\Rightarrow$'': Let $M$ be a FLP-answer set of $P$.
We will show that $M$ is an answer set of $P$ (w.r.t. 
Definition \ref{basic-pos-a.s.}). 

Let $Q = FLP(P,M)$.  From the definition of 
FLP-answer set, $M$ is a minimal model of $Q$. 
Let $M' = T^\infty_{P}(\emptyset, M)$. 
As $M$ is a model of $Q$, it is also a model of $P$. 
Corollary \ref{cr-tp} implies that $M' \subseteq M$. 

Consider $r \in Q$ such that $M' \models body(r)$
and $head(r) = (\{a\},\{\{a\}\})$. From the definition of
$Q$ and the monotonicity of $P$, we have $M \models body(r)$. 
It follows from Lemma \ref{lm5} that $M' \models_M body(r)$. Hence,
$a \in M'$ (by the definition of the operator $T_P$).
This implies that $M'$ is a model of $Q$. 

Because of the minimality of $M$ and $M' \subseteq M$,
we have $M' = M$. Hence, $M$ is an answer set of $P$.

\item ``$\Leftarrow$'': Let $M$ be an answer set of $P$.
We will prove that $M$ is a FLP-answer set of $P$ by 
showing that $M$ is a minimal model of $Q=FLP(P,M)$. 

Let $M' \subseteq M$ be a model of $Q$.
First, we will demonstrate that $M'$ is a model of
$P$. Suppose otherwise, i.e., $M'$ is not a model
of $P$. This implies that $P$ contains a rule $r$
such that 
$head(r) = (\{a\},\{\{a\}\})$ for some atom $a$,
$M' \models A$ for $A \in pos(r)$, and  $a \not \in M'$.
Due to the monotonicity of $P$ we have that 
$M \models A$ for $A \in pos(r)$. Hence, $Q$ contains the 
rule $r$. As a result, we have $M' \models body(r)$. Thus,
$a \in M'$ because $M'$ is a model of $Q$. This is a contradiction. 

We have shown that $M'$ is a model of $P$. On the other 
hand, by Corollary \ref{cr-minm}, $M$ is a minimal model of $P$.
Therefore, we have $M \subseteq M'$. Accordingly, we have
$M' = M$. Thus, $M$ is a minimal model of $Q$, i.e., an FLP-answer set of $P$.
\end{enumerate}
\hfill$\Box$

\bigskip 
\noindent
{\bf Proof of Proposition~\ref{equiv-pelov}.}
Let $P$ be a positive program with aggregates and $R$ 
and $M$ be two set of atoms such that $R \subseteq M$. Then, 
$T_P(R,M) = \Phi^{1}(R,M)$ 
where $\Phi^{aggr}(R,M) = (\Phi^{1}(R,M), \Phi^{2}(R,M))$. 

\smallskip
\noindent
{\bf Proof.}
In order to prove this result, we will make use of an intermediate
step. In~\citeA{SonP05}, the following concepts for program with
aggregates are introduced:
\begin{list}{$\circ$}{\topsep=1pt \parsep=0pt \itemsep=1pt}
\item Given an aggregate $A$, a solution of $A$ is a pair
    $\langle S^+,S^-\rangle$, satisfying the following 
    properties:
    \begin{list}{$*$}{\topsep=1pt \parsep=0pt \itemsep=1pt}
    \item  $S^+\subseteq {\cal A}$ and $S^-\subseteq {\cal A}$,
    \item  $S^+ \cap S^- = \emptyset$, and 
    \item for each $I \subseteq {\cal A}$ where $S^+\subseteq I$
    and $I\cap S^-=\emptyset$, we have that $I \models A$.
    \end{list}
\item Given two interpretations $I,M$, an aggregate $A$ is 
    conditionally satisfied w.r.t. $I,M$ (denoted $(I,M)\models A$) if 
    $\langle I\cap M \cap A_d, A_d \setminus M\rangle$ is a solution
    of $A$. For simplicity, we define also conditional satisfaction for
    atoms, by saying that $a$ is conditionally satisfied w.r.t. $I,M$ if
    $a \in I$.
\item Given a positive program with aggregates $P$ and an interpretation $M$,
    the aggregate consequence operator $K^P_M: 2^{\cal A} \rightarrow 2^{\cal A}$ 
    is defined as:\footnote{The original definition in \citeA{SonP05} allows for the use of
    negative atoms in the body of the rules, but we omit this for the sake
    of simplicity.} $K^P_M(I) = \{head(r) |r \in P, (I,M)\models body(r)\}$.
\end{list}
We wish to show here that, for a positive program with aggregates $P$ and
for interpretations $I,M$, $K^P_M(I) = T_P(I,M)$. This will allow us to conclude
the result of proposition~\ref{equiv-pelov}, since it has been proved that 
$K^P_M(I) = \Phi^1(I,M)$~\cite{SonP05}.

Observe that, under the condition $I\subseteq M$:
\begin{itemize}
\item If $a$ is a standard atom, then $I\models_M a$ iff $a\in I$ iff $(I,M)\models a$.
\item Let $A$ be  an aggregate.
    \begin{itemize}
    \item Let us assume $I\models_M A$. This means that $I\models A$ and, for each
        $J\leq A_d$ s.t. $I \cap A_d \subseteq J \subseteq M\cap A_d$, we have
        that $J\models A$.  
    
        If we consider $J\subseteq A_d$ s.t. $I\cap M \cap A_d \subseteq J$ and 
        $J\cap (A_d\setminus M)=\emptyset$, then $I\cap M\cap A_d = I\cap A_d \subseteq J$, and
        $J\subseteq M\cap A_d$ (otherwise, if $a \in J$ and $a \not\in M\cap A_d$, then
        $a \in A_d \setminus M$, which would violate the condition
        $J\cap (A_d\setminus M)=\emptyset$). From the initial assumption that
        $I\models_M A$, we can conclude that $J\in A_c$. This allows us to
        conclude that $I\models_M A$ implies $(I,M)\models A$.

    \item Let us assume $(I,M)\models A$. This means that, for each $J\subseteq A_d$ 
        s.t. $I\cap M\cap A_d \subseteq J$ and $J\cap (A_d\setminus M)=\emptyset$, we
        have that $J\in A_c$.

        First of all, note that $I\cap A_d = I\cap M \cap A_d$, thus 
        $I\cap A_d \in A_c$---i.e., $I\models A$.
        Let us now take some arbitrary $J\subseteq A_d$, where $I\cap A_d \subseteq J\subseteq M \cap A_d$.
        Since $I\cap A_d\subseteq J$, in particular $I \cap M \cap A_d \subseteq J$. Furthermore,
        $J\cap (A_d\setminus M) = \emptyset$, since $J \subseteq M\cap A_d$. Thus, from the initial
        assumption, we have $J\models A_c$. This allows us to conclude that
        $(I,M)\models A$ implies $I\models_M A$. 
                
    \end{itemize}
\end{itemize}
These results allows us to conclude that for any element $\alpha$ in the body of a rule of $P$
(either atom or aggregate), $(I,M)\models \alpha$ iff $I\models_M \alpha$. This allows us to 
immediately conclude that $K^P_M(I) = T_P(I,M)$.
\qed

\bibliographystyle{theapa}

\begin{thebibliography}{99}

\bibitem[\protect\BCAY{Banti, Alferes, Brogi,\ \BBA\ Hitzler}{Banti
  et~al.}{2005}]{BantiABH05}
Banti, F., Alferes, J.~J., Brogi, A., \BBA\ Hitzler, P. \BBOP2005\BBCP.
\newblock \BBOQ The well supported semantics for multidimensional dynamic logic
  programs.\BBCQ\
\newblock In Baral, C., Greco, G., Leone, N., \BBA\ Terracina, G.\BEDS, {\Bem
  Logic Programming and Nonmonotonic Reasoning, 8th International Conference,
  LPNMR 2005, Diamante, Italy, September 5-8, 2005, Proceedings},
  \lowercase{\BVOL}\ 3662 of {\Bem Lecture Notes in Computer Science}, \BPGS\
  356--368. Springer.

\bibitem[\protect\BCAY{Baral}{Baral}{2003}]{Baral03}
Baral, C. \BBOP{2003}\BBCP.
\newblock {\Bem {K}nowledge {R}epresentation, reasoning, and declarative
  problem solving with {A}nswer sets}.
\newblock {Cambridge University Press, Cambridge, MA}.

\bibitem[\protect\BCAY{Baral}{Baral}{2005}]{Baral05}
Baral, C. \BBOP2005\BBCP.
\newblock \BBOQ {From Knowledge to Intelligence --- Building Blocks and
  Applications}\BBCQ.
\newblock {Invited Talk, AAAI,
  \url{www.public.asu.edu/~cbaral/aaai05-invited-talk.ppt}}.

\bibitem[\protect\BCAY{Dell'Armi, Faber, Ielpa, Leone,\ \BBA\
  Pfeifer}{Dell'Armi et~al.}{2003}]{Dell-ArmiFILP03}
Dell'Armi, T., Faber, W., Ielpa, G., Leone, N., \BBA\ Pfeifer, G.
  \BBOP2003\BBCP.
\newblock \BBOQ {Aggregate Functions in Disjunctive Logic Programming:
  Semantics,Complexity,and Implementation in DLV}\BBCQ\
\newblock In {\Bem {Proceedings of the 18th International Joint Conference on
  Artificial Intelligence (IJCAI) 2003}}, \BPGS\ 847--852.

\bibitem[\protect\BCAY{Denecker, Pelov,\ \BBA\ Bruynooghe}{Denecker
  et~al.}{2001}]{DeneckerPB01}
Denecker, M., Pelov, N., \BBA\ Bruynooghe, M. \BBOP2001\BBCP.
\newblock \BBOQ Ultimate well-founded and stable semantics for logic programs
  with aggregates.\BBCQ\
\newblock In Codognet, P.\BED, {\Bem Logic Programming, 17th International
  Conference, ICLP 2001, Paphos, Cyprus, November 26 - December 1, 2001,
  Proceedings}, \lowercase{\BVOL}\ 2237 of {\Bem Lecture Notes in Computer
  Science}, \BPGS\ 212--226. Springer.

\bibitem[\protect\BCAY{Eiter, Leone, Mateis, Pfeifer,\ \BBA\ Scarcello}{Eiter
  et~al.}{1998}]{eiter98a}
Eiter, T., Leone, N., Mateis, C., Pfeifer, G., \BBA\ Scarcello, F.
  \BBOP1998\BBCP.
\newblock \BBOQ {The KR System {\tt dlv}: Progress Report, Comparisons, and
  Benchmarks}\BBCQ\
\newblock In {\Bem International Conference on Principles of Knowledge
  Representation and Reasoning}, \BPGS\ 406--417.

\bibitem[\protect\BCAY{Elkabani, Pontelli,\ \BBA\ Son}{Elkabani
  et~al.}{2004}]{ElkabaniPS04}
Elkabani, I., Pontelli, E., \BBA\ Son, T.~C. \BBOP2004\BBCP.
\newblock \BBOQ Smodels with CLP and its applications: A simple and effective
  approach to aggregates in asp.\BBCQ\
\newblock In Demoen, B.\BBACOMMA\  \BBA\ Lifschitz, V.\BEDS, {\Bem Logic
  Programming, 20th International Conference, ICLP 2004, Saint-Malo, France,
  September 6-10, 2004, Proceedings}, \lowercase{\BVOL}\ 3132 of {\Bem Lecture
  Notes in Computer Science}, \BPGS\ 73--89. Springer.

\bibitem[\protect\BCAY{Elkabani, Pontelli,\ \BBA\ Son}{Elkabani
  et~al.}{2005}]{ElkabaniPS05}
Elkabani, I., Pontelli, E., \BBA\ Son, T.~C. \BBOP2005\BBCP.
\newblock \BBOQ {Smodels$^{\mbox{A}}$ - A System for Computing Answer Sets of
  Logic Programs with Aggregates.}\BBCQ\
\newblock In Baral, C., Greco, G., Leone, N., \BBA\ Terracina, G.\BEDS, {\Bem
  Logic Programming and Nonmonotonic Reasoning, 8th International Conference,
  LPNMR 2005, Diamante, Italy, September 5-8, 2005, Proceedings},
  \lowercase{\BVOL}\ 3662 of {\Bem Lecture Notes in Computer Science}, \BPGS\
  427--431. Springer.

\bibitem[\protect\BCAY{Faber, Leone,\ \BBA\ Pfeifer}{Faber
  et~al.}{2004}]{FaberLP04}
Faber, W., Leone, N., \BBA\ Pfeifer, G. \BBOP2004\BBCP.
\newblock \BBOQ Recursive aggregates in disjunctive logic programs: Semantics
  and complexity.\BBCQ\
\newblock In Alferes, J.~J. \BBACOMMA\  \BBA\ Leite, J.~A.\BEDS, {\Bem 
   Logics in Artificial Intelligence, 9th European Conference,
   JELIA 2004, Lisbon, Portugal, September 27-30, 2004, Proceedings},
   \lowercase{\BVOL}\ 3229 of {\Bem Lecture
  Notes in Computer Science}, \BPGS\ 200--212. Springer.

\bibitem[\protect\BCAY{Fages}{Fages}{1994}]{Fages94}
Fages, F. \BBOP1994\BBCP.
\newblock \BBOQ Consistency of {C}lark's completion and existence of stable
  models\BBCQ\
\newblock {\Bem Methods of Logic in Computer Science}, \BPGS\ 51--60.

\bibitem[\protect\BCAY{Ferraris}{Ferraris}{2005}]{Ferraris05a}
Ferraris, P. \BBOP2005\BBCP.
\newblock \BBOQ Answer sets for propositional theories.\BBCQ\
\newblock In Baral, C., Greco, G., Leone, N., \BBA\ Terracina, G.\BEDS, {\Bem
  Logic Programming and Nonmonotonic Reasoning, 8th International Conference,
  LPNMR 2005, Diamante, Italy, September 5-8, 2005, Proceedings},
  \lowercase{\BVOL}\ 3662 of {\Bem Lecture Notes in Computer Science}, \BPGS\
  119--131. Springer.

\bibitem[\protect\BCAY{Gelder}{Gelder}{1992}]{Gelder92}
Gelder, A.~V. \BBOP1992\BBCP.
\newblock \BBOQ The well-founded semantics of aggregation.\BBCQ\
\newblock In {\Bem Proceedings of the Eleventh ACM SIGACT-SIGMOD-SIGART
  Symposium on Principles of Database Systems, June 2-4, 1992, San Diego,
  California}, \BPGS\ 127--138. ACM Press.

\bibitem[\protect\BCAY{Gelfond\ \BBA\ Leone}{Gelfond\ \BBA\
  Leone}{2002}]{GelfondL02}
Gelfond, M.\BBACOMMA\  \BBA\ Leone, N. \BBOP2002\BBCP.
\newblock \BBOQ Logic programming and knowledge representation -- the
  {A-Prolog} perspective\BBCQ\
\newblock {\Bem Artificial Intelligence}, {\Bem 138\/}(1-2), 3--38.

\bibitem[\protect\BCAY{Gelfond\ \BBA\ Lifschitz}{Gelfond\ \BBA\
  Lifschitz}{1988}]{GelfondL88}
Gelfond, M.\BBACOMMA\  \BBA\ Lifschitz, V. \BBOP1988\BBCP.
\newblock \BBOQ The stable model semantics for logic programming\BBCQ\
\newblock In Kowalski, R.\BBACOMMA\  \BBA\ Bowen, K.\BEDS, {\Bem Logic
  Programming: Proceedings~of the Fifth International Conf.~and Symp.}, \BPGS\
  1070--1080.

\bibitem[\protect\BCAY{Gelfond}{Gelfond}{2002}]{a-prolog}
Gelfond, M. \BBOP2002\BBCP.
\newblock \BBOQ {Representing Knowledge in A-Prolog}\BBCQ\
\newblock In Kakas, A.\BBACOMMA\  \BBA\ Sadri, F.\BEDS, {\Bem Computational
  Logic: Logic Programming and Beyond}, \BPGS\ 413--451. Springer Verlag.

\bibitem[\protect\BCAY{Hitzler\ \BBA\ Wendt}{Hitzler\ \BBA\
  Wendt}{2005}]{HitzlerW05}
Hitzler, P.\BBACOMMA\  \BBA\ Wendt, M. \BBOP2005\BBCP.
\newblock \BBOQ A uniform approach to logic programming semantics\BBCQ\
\newblock {\Bem {Theory and Practice of Logic Programming}}, {\Bem 5\/}(1-2),
  123--159.

\bibitem[\protect\BCAY{Kemp\ \BBA\ Stuckey}{Kemp\ \BBA\
  Stuckey}{1991}]{KempS91}
Kemp, D.~B.\BBACOMMA\  \BBA\ Stuckey, P.~J. \BBOP1991\BBCP.
\newblock \BBOQ Semantics of logic programs with aggregates.\BBCQ\
\newblock In Saraswat, V.~A  \BBACOMMA\  \BBA\ Ueda, K. 
\BEDS, {\Bem Logic Programming, Proceedings of the 1991 International 
Symposium, San Diego, California, USA}, 
\BPGS\ 387-401. MIT Press.

\bibitem[\protect\BCAY{Liu\ \BBA\ Truszczy\'{n}ski}{Liu\ \BBA\
  Truszczynski}{2005a}]{LiuT05b}
Liu, L.\BBACOMMA\  \BBA\ Truszczy\'{n}ski, M. \BBOP2005a\BBCP.
\newblock \BBOQ Pbmodels - software to compute stable models by pseudoboolean
  solvers.\BBCQ\
\newblock In Baral, C., Greco, G., Leone, N., \BBA\ Terracina, G.\BEDS, {\Bem
  Logic Programming and Nonmonotonic Reasoning, 8th International Conference,
  LPNMR 2005, Diamante, Italy, September 5-8, 2005, Proceedings},
  \lowercase{\BVOL}\ 3662 of {\Bem Lecture Notes in Computer Science}, \BPGS\
  410--415.

\bibitem[\protect\BCAY{Liu\ \BBA\ Truszczy\'{n}ski}{Liu\ \BBA\
  Truszczy\'{n}ski}{2005b}]{LiuT05}
Liu, L.\BBACOMMA\  \BBA\ Truszczy\'{n}ski, M. \BBOP2005b\BBCP.
\newblock \BBOQ Properties of programs with monotone and convex
  constraints.\BBCQ\
\newblock In Veloso, M.~M.\BBACOMMA\  \BBA\ Kambhampati, S.\BEDS, {\Bem
  Proceedings, The Twentieth National Conference on Artificial Intelligence and
  the Seventeenth Innovative Applications of Artificial Intelligence
  Conference, July 9-13, 2005, Pittsburgh, Pennsylvania, USA}, \BPGS\ 701--706.
  AAAI Press AAAI Press / The MIT Press.

\bibitem[\protect\BCAY{Marek\ \BBA\ Remmel}{Marek\ \BBA\
  Remmel}{2004}]{MarekR04}
Marek, V.~W.\BBACOMMA\  \BBA\ Remmel, J.~B. \BBOP2004\BBCP.
\newblock \BBOQ Set constraints in logic programming\BBCQ\
\newblock In {\Bem Logic Programming and Nonmonotonic Reasoning, 7th
  International Conference, LPNMR 2004, Fort Lauderdale, FL, USA, January 6-8,
  2004, Proceedings}, \lowercase{\BVOL}\ 2923 of {\Bem Lecture Notes in
  Computer Science}, \BPGS\ 167--179. Springer Verlag.

\bibitem[\protect\BCAY{Marek\ \BBA\ Truszczy\'{n}ski}{Marek\ \BBA\
  Truszczy\'{n}ski}{1999}]{MarekT99}
Marek, V.~W.\BBACOMMA\  \BBA\ Truszczy\'{n}ski, M. \BBOP1999\BBCP.
\newblock \BBOQ Stable Models as an Alternative Logic Programming
	Paradigm.\BBCQ\
\newblock In {\Bem The Logic Programming Paradigm}, 
	Springer Verlag.

\bibitem[\protect\BCAY{Marek\ \BBA\ Truszczy\'{n}ski}{Marek\ \BBA\
  Truszczy\'{n}ski}{2004}]{MarekT04}
Marek, V.~W.\BBACOMMA\  \BBA\ Truszczy\'{n}ski, M. \BBOP2004\BBCP.
\newblock \BBOQ Logic programs with abstract constraint atoms.\BBCQ\
\newblock In {\Bem Proceedings of the Nineteenth National Conference on
  Artificial Intelligence, Sixteenth Conference on Innovative Applications of
  Artificial Intelligence, July 25-29, 2004, San Jose, California, USA}. AAAI
  Press / The MIT Press.

\bibitem[\protect\BCAY{Mumick, Pirahesh,\ \BBA\ Ramakrishnan}{Mumick
  et~al.}{1990}]{MumickPR90}
Mumick, I.~S., Pirahesh, H., \BBA\ Ramakrishnan, R. \BBOP1990\BBCP.
\newblock \BBOQ The magic of duplicates and aggregates.\BBCQ\
\newblock In McLeod, D., Sacks-Davis, R., \BBA\ Schek, H.-J.\BEDS, {\Bem 16th
  International Conference on Very Large Data Bases, August 13-16, 1990,
  Brisbane, Queensland, Australia, Proceeding}, \BPGS\ 264--277. Morgan
  Kaufmann.

\bibitem[\protect\BCAY{Niemel{\"{a}}}{Niemel{\"{a}}}{1999}]{Niemela99}
Niemel{\"{a}}, I.,  \BBOP1999\BBCP.
\newblock \BBOQ Logic Programs with Stable Models as a Constraint Programming Paradigm.\BBCQ\
\newblock In {\Bem Annals of Math and AI}, {\Bem 25\/}(3--4), 241--273.

\bibitem[\protect\BCAY{Niemel{\"{a}}, Simons,\ \BBA\ Soininen}{Niemel{\"{a}}
  et~al.}{1999}]{nie99b}
Niemel{\"{a}}, I., Simons, P., \BBA\ Soininen, T. \BBOP1999\BBCP.
\newblock \BBOQ Stable model semantics for weight constraint rules\BBCQ\
\newblock In {\Bem Proceedings of the 5th International Conference on on Logic
  Programming and Nonmonotonic Reasoning}, \BPGS\ 315--332.

\bibitem[\protect\BCAY{Pelov}{Pelov}{2004}]{Pelov04}
Pelov, N. \BBOP2004\BBCP.
\newblock {\Bem {Semantic of Logic Programs with Aggregates}}.
\newblock Ph.D.\ thesis, Katholieke Universiteit Leuven.
\newblock
  \url{http://www.cs.kuleuven.ac.be/publicaties/doctoraten/cw/CW2004_02.abs.html}.

\bibitem[\protect\BCAY{Pelov\ \BBA\ Truszczy\'{n}ski}{Pelov\ \BBA\
  Truszczy\'{n}ski}{2004}]{PelovT04}
Pelov, N.\BBACOMMA\  \BBA\ Truszczy\'{n}ski, M. \BBOP2004\BBCP.
\newblock \BBOQ Semantics of disjunctive programs with monotone aggregates ---
  an operator-based approach.\BBCQ\
\newblock In Delgrande, J.~P.\BBACOMMA\  \BBA\ Schaub, T.\BEDS, {\Bem 10th
  International Workshop on Non-Monotonic Reasoning (NMR 2004), Whistler,
  Canada, June 6-8, 2004, Proceedings}, \BPGS\ 327--334.

\bibitem[\protect\BCAY{Simons, Niemel{\"{a}},\ \BBA\ Soininen}{Simons
  et~al.}{2002}]{sim02}
Simons, P., Niemel{\"{a}}, N., \BBA\ Soininen, T. \BBOP2002\BBCP.
\newblock \BBOQ {Extending and Implementing the Stable Model Semantics}\BBCQ\
\newblock {\Bem Artificial Intelligence}, {\Bem 138\/}(1--2), 181--234.

\bibitem[\protect\BCAY{Son\ \BBA\ Pontelli}{Son\ \BBA\ Pontelli}{2007}]{SonP05}
Son, T.~C.\BBACOMMA\  \BBA\ Pontelli, E. \BBOP2007\BBCP.
\newblock \BBOQ {A Constructive Semantic Characterization of Aggregates in
  Answer Set Programming}\BBCQ\
\newblock {\Bem {Theory and Practice of Logic Programming}}.
\newblock {\Bem 7\/}(03), 355--375. 

\bibitem[\protect\BCAY{Son, Pontelli,\ \BBA\ Elkabani}{Son, Pontelli,\ 
\BBA\ Elkabani}{2006}]{SonPE06}
Son, T.~C., Pontelli, E., \BBA\ Elkabani, I. \BBOP2006\BBCP.
\newblock \BBOQ {An Unfolding-Based Semantics for Logic Programming with
  Aggregates}\BBCQ\
\newblock {\Bem {Computing Research Repository}}.
\newblock {cs.SE/0605038}.

\end{thebibliography}

\end{document}